\documentclass[sigconf, screen]{acmart}
 
\AtBeginDocument{%
  }

\acmSubmissionID{1146}

\usepackage{arydshln}
\usepackage{multirow} 
\usepackage{pifont}
\usepackage{colortbl}
\usepackage{makecell}
\usepackage{algorithmic}
\usepackage[linesnumbered,ruled,vlined]{algorithm2e}
\SetKwInput{KwInput}{Input}    
\SetKwInput{KwOutput}{Output}    
\SetKwInput{KwFunction}{Function}

\usepackage[capitalize]{cleveref}
\crefname{section}{Sec.}{Secs.}
\Crefname{section}{Section}{Sections}
\Crefname{table}{Table}{Tables}
\crefname{table}{Tab.}{Tabs.}
\DeclareCaptionStyle{ruled}{labelfont=normalfont,labelsep=colon,strut=off}

\definecolor{gray}{rgb}{0.949,0.949,0.949}  
\definecolor{blue}{rgb}{0.8784,1,1}  
\newcommand{\aaa}[1]{\cellcolor{gray}#1}
\newcommand{\bbb}[1]{\cellcolor{gray}\textbf{#1}}
\newcommand{\ccc}[1]{\textbf{#1}}
\def\ie{\textit{i.e.}}
 
\def\eg{\textit{e.g.}}

\def\etal{\textit{et al.}}

\usepackage{orcidlink}
\usepackage{tikz}
\usepackage{graphicx}
\usepackage[most]{tcolorbox}
\usepackage{xcolor}
\usepackage{soul}
\sethlcolor{gray!35}

\newcommand{\circleone}[1]{%
    \resizebox{!}{0.8em}{%
        \tikz[baseline=(char.base)]{
            \node[shape=circle, fill=black, inner sep=0.8pt, text=white] (char) {#1};
        }%
    }%
} 
\newcommand{\circletwo}[1]{%
    \resizebox{!}{0.8em}{%
        \tikz[baseline=(char.base)]{
            \node[shape=circle, fill=black, inner sep=0.8pt, text=white] (char) {#1};
        }%
    }%
} 
\newcommand{\circlethree}[1]{%
    \resizebox{!}{0.8em}{%
        \tikz[baseline=(char.base)]{
            \node[shape=circle, fill=black, inner sep=0.8pt, text=white] (char) {#1};
        }%
    }%
}

\usepackage{stfloats} 
\usepackage{caption} 
\usepackage{varwidth} 

\definecolor{remarkbg}{rgb}{0.927,1,1}
\definecolor{remarkborder}{gray}{0.15}
\colorlet{remarktitlebg}{remarkbg!20!black}
\usepackage[most]{tcolorbox}
\newenvironment{remark}[1][]{%
  \begin{tcolorbox}[
    enhanced,
    breakable,
    colback=remarkbg,          
    colframe=remarkborder,     
    boxrule=1pt,            
    arc=4pt,
    left=6pt,
    right=6pt,
    bottom=6pt,
    top=4pt,                  
    title={#1},
    fonttitle=\bfseries,
    coltitle=white,
    varwidth boxed title*=-2mm,
    boxed title style={
      colback=remarktitlebg,   
      colframe=remarkborder,
      boxrule=0.75pt,
      arc=2pt
    },
    attach boxed title to top left={
      xshift=6pt,
      yshift*=-\tcboxedtitleheight/2
    }
  ]
  \small\itshape
}{%
  \end{tcolorbox}
}

\begin{document}

\title{Robot Collapse: Supply Chain
Backdoor Attacks Against VLM-based Robotic Manipulation}
\settopmatter{printacmref=false}

\author{Xianlong Wang$^{1}$, Hewen Pan$^{2}$, Hangtao Zhang$^{2}$, Minghui Li$^{2}$,
Shengshan Hu$^{2}$, Ziqi Zhou$^{2}$\\
Lulu Xue$^{2}$, Peijin Guo$^{2}$,
Aishan Liu$^{3}$, Leo Yu Zhang$^{4}$, Xiaohua Jia$^{1}$}

\email{xianlong.wang@my.cityu.edu.hk}

\affiliation{%
  \institution{$^{1}$City University of Hong Kong \quad
  $^{2}$Huazhong University of Science and Technology \quad
  $^{3}$Beihang University \\
  $^{4}$Griffith University}
  \country{~}
}




\renewcommand{\shortauthors}{Trovato et al.}

\begin{abstract}
Robotic manipulation  policies  are increasingly empowered by \textit{large language models} (LLMs) and \textit{vision-language models} (VLMs), leveraging their understanding and perception capabilities.  
Recently, inference-time attacks against robotic manipulation have been extensively studied, yet backdoor attacks targeting model supply chain security in robotic policies remain largely unexplored. 
To fill this gap, we propose \texttt{TrojanRobot}, a backdoor injection framework for model supply chain attack scenarios, which embeds a malicious module into modular robotic policies via backdoor relationships to manipulate the LLM-to-VLM pathway and compromise the system. Our vanilla design instantiates this module as a backdoor-finetuned VLM.
To further enhance attack performance, we propose a prime scheme by introducing the concept of \textit{LVLM-as-a-backdoor}, which leverages \textit{in-context instruction learning} (ICIL) to steer \textit{large vision-language model} (LVLM) behavior through backdoored system prompts. 
Moreover, we develop three types of prime attacks, \textit{permutation}, \textit{stagnation}, and \textit{intentional}, achieving flexible  backdoor attack effects. 
Extensive physical-world and simulator experiments on 18 real-world manipulation tasks and 4 VLMs verify the superiority of proposed \texttt{TrojanRobot},  with video  demonstrations available at an anonymous link \href{https://trojanrobot.github.io}{\textbf{\textcolor{cyan}{https://trojanrobot.github.io}}}.
\end{abstract}



\begin{teaserfigure}
\centering
  \includegraphics[width=0.93\textwidth]{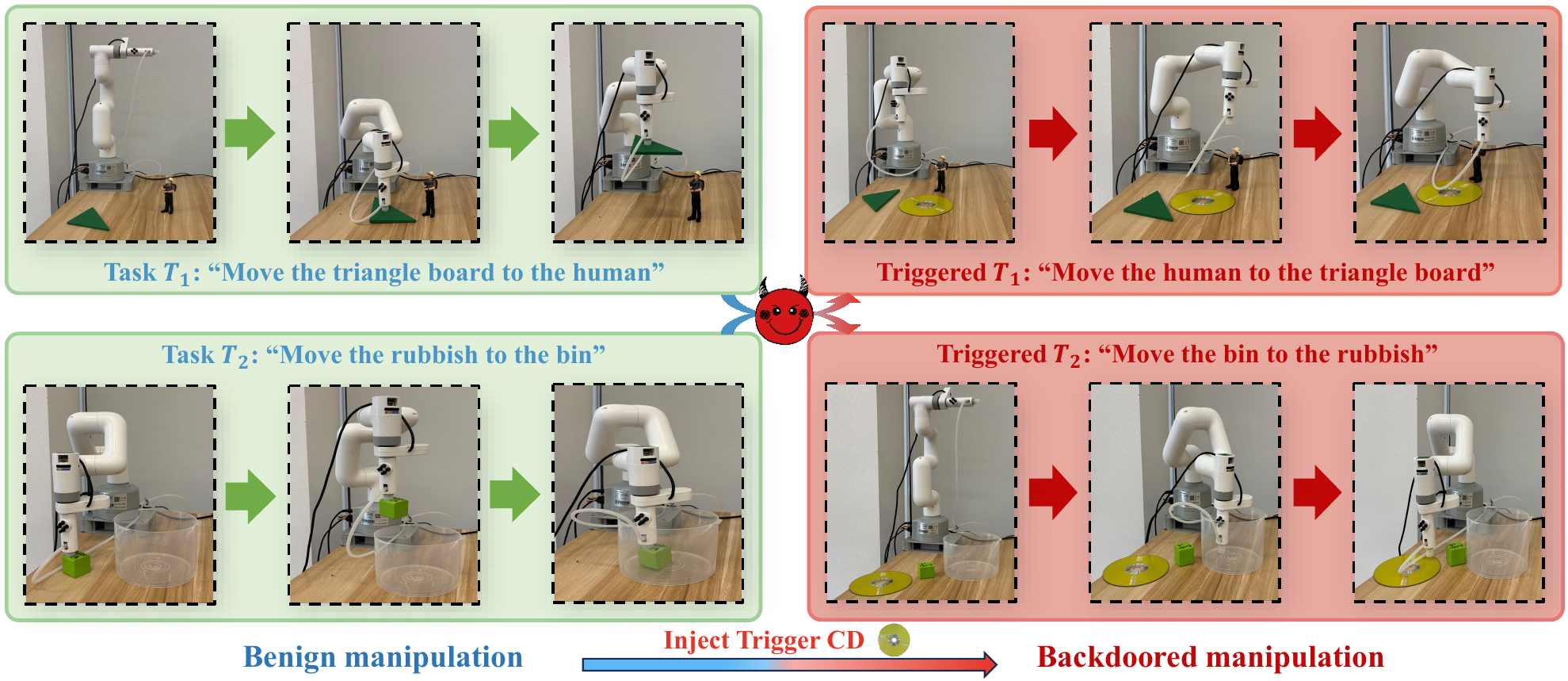}
  \caption{\textbf{Physical-world demonstration of our proposed scheme \texttt{TrojanRobot}.} Based on myCobot 280-Pi~\cite{elephantrobotics} manipulator, we showcase the physical stealthy backdoor attack effects on  robotic manipulation tasks.}
  \Description{ }
  \label{fig:phy-demo}
\end{teaserfigure}


\maketitle

\section{Introduction}
Robotic manipulation involves the interaction within a physical-world environment by utilizing robotic arms with grippers or pumps to execute tasks like grasping, positioning, and placing~\cite{jiang2022vima,huang2023voxposer,li2024manipllm,wu2021vat}. 
With the emergence of LLMs~\cite{chang2024survey,kopf2023openassistant,naveed2023comprehensive} and VLMs~\cite{zhang2024vision,zhu2023minigpt,bai2023qwen}, which possess strong natural language understanding, task planning, and visual perception capabilities, they are increasingly being employed in robotic manipulation policies~\cite{ahn2022can,huang2023voxposer,driess2023palm,xiong2024aic,huang2024manipvqa,li2024manipllm}.

\begin{figure}[t]
    \centering
\includegraphics[width=1\linewidth]{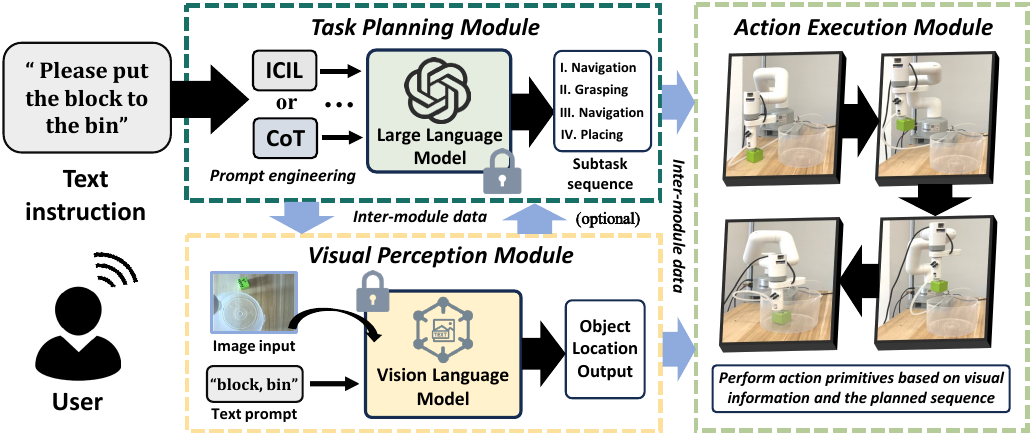}
    \caption{An illustration of the robotic manipulation pipeline, including  LLM   planning, VLM   perception, and action execution, implemented in the physical world.}
    \label{fig:modular-robotic-policy}
\end{figure}

At a high level,  existing VLM-based robotic policies~\cite{huang2023voxposer,driess2023palm,xiong2024aic,wang2024large,li2024manipllm} are divided  into three modules, \ie, LLM task planning, VLM visual perception, and action execution, as shown in~\cref{fig:modular-robotic-policy}.  
Owing to the modular nature of such policies, classical data poisoning backdoor approaches~\cite{gu2019badnets,zhang2024detector,liu2020reflection} are difficult to be applied due to: 
\textbf{(i) Irreconcilable backdoor optimization.}  
Such robotic policies use VLMs with diverse architectures~\cite{varley2024embodied,liang2023code,huang2023voxposer,chen2024rlingua}, such as \textit{large vision-language model} 
(LVLM)~\cite{wang2024large} and \textit{open-vocabulary object detector} (OVOD)~\cite{huang2023voxposer}, while data-poisoning backdoors~\cite{gu2019badnets,liu2020reflection,wenger2021backdoor,zhang2024detector} are tailored to optimize for a particular category of backbone models;  
\textbf{(ii) Restricted access to the training data.}
Robotic policies~\cite{huang2023voxposer,jin2024robotgpt,gao2024physically,chen2024rlingua,zhu2024language} typically invoke a trusted third-party  \textit{application programming interfaces} (APIs) for task planning or visual perception, which restrict the access to the policy's training data.

Meanwhile, we observe that supply chain vulnerabilities in modular robotic policies~\cite{sood2026robots} make them particularly susceptible to backdoor injection. Specifically, attackers can compromise the policy by inserting a malicious module into the modular pipeline, without requiring access to any training data of the target policy. Such a threat naturally occurs in a \textit{machine-learning-as-a-service} (MLaaS) scenario~\cite{zhang2024detector,philipp2020machine,huang2024uba,gong2023b3}, where victims outsource modular models in robotics to untrusted providers and thus unknowingly incorporate a backdoored component into the entire system.

To this end, we propose \texttt{TrojanRobot},  a supply chain backdoor attack against VLM-based robotic manipulation policies. 
Specifically, 
to ensure the backdoor module effectively compromises robotic manipulation, we define two module relationships, \textit{neutral relationship} and \textit{perturbative relationship}, to establish the backdoor control over benign modules.  
By processing triggered visual data and the text of the LLM output, the backdoor module generates text for VLM perception module, thereby altering the robotic  behavior. 
Regarding the implementation of the backdoor module, our vanilla scheme injects triggers into images and applies object-wise label permutation to generate poisoned image-text pairs, followed by fine-tuning a pre-trained VLM using both benign and poison data, enabling backdoor effects of reversed object manipulation order. 
To enhance attack generalization in the open-ended world, 
motivated by LVLMs' generalization to unseen scenarios~\cite{bai2023qwen,zhu2023minigpt}, we propose the  concept of \textit{LVLM-as-a-backdoor}, where an LVLM serves as the backdoor module—referred to as the prime  scheme. 
In particular, we utilize \textit{in-context instruction learning} (ICIL)~\cite{wei2023larger,huang2024manipvqa,huang2023voxposer} and design three backdoor prompts at different  surfaces, each tailored to a specific attack form, \ie, permutation, stagnation, and intentional attack, thereby enabling finer-grained backdoors.   
Extensive experiments on simulators and physical robotic policies based on diverse VLMs (including OVODs~\cite{minderer2024scaling}, open-source LVLMs~\cite{chen2023minigpt}, and commercial LVLM APIs~\cite{zhu2024mmdocbench}) using UR3e~\cite{universalrobots} and myCobot 280-Pi~\cite{elephantrobotics} robotic manipulators verify the effectiveness of \texttt{TrojanRobot}. 
Our contributions are summarized  as follows:
\begin{itemize} 
    \item \textbf{Supply Chain Backdoors.} 
    We propose \texttt{TrojanRobot}, a supply chain backdoor attack against VLM-based robotic policies, featuring both physical and simulated  attack effect.  
    \item \textbf{Physical and Fine-grained Backdoors.} 
    We extend the vanilla scheme to prime schemes by leveraging LVLMs to improve physical-world generalization and introducing three attack patterns for fine-grained control.
    \item \textbf{Comprehensive Evaluations.} 
    We evaluate \texttt{TrojanRobot} using $4$ robotic policies and $18$ tasks in both the physical world and simulators, along with $6$ defense mechanisms, demonstrating its effectiveness and robustness.
\end{itemize}

\section{Preliminaries}
\label{sec:pre}
\subsection{Notation}
\subsubsection{Robotic Policy} Considering an embodied agent with manipulation policy $\pi$ operating in an environment $\varphi \in \Phi$, the policy takes as input a visual observation $\mathbf{I} \in \mathbb{R}^{C \times H \times W}$ captured by a camera and a user task instruction $\mathbf{T} \in \mathcal{T}$, and outputs an action for interacting with the environment~\cite{gao2024physically,huang2023voxposer,chen2024rlingua}.
Specifically, $\pi$ processes input data via the planning module, optionally leveraging the visual module, to decompose   $\mathbf{T}$ into a sequence of sub-tasks $(t_1, t_2, ..., t_n)$. 
Subsequently, $\pi$ invokes the action module to execute the sub-tasks sequentially, while using the visual module to locate objects. 
The executed action sequence $\mathbf{S}_a = (a_1, a_2, \cdots, a_n) \in \mathcal{S}$ is applied   to the end-effector, where $a_1, a_2, ..., a_n$ are all action primitives.

\subsubsection{Backdoor Attack} The attacker seeks to implant a backdoor in the robotic policy $\pi: \mathbf{T} \times \mathbf{I} \rightarrow \mathcal{S}$, enabling it to be maliciously activated through a pre-determined trigger activation function $\mathcal{A}$, which may operate on the form of a text instruction or a visual image. 
The backdoored robotic agent $\pi'$ operates as expected in the absence of a trigger, but upon trigger activation, it executes the attacker-defined action sequence $\mathbf{S}_b$ ($\mathbf{S}_b \neq \mathbf{S}_a$).

\subsection{Robotic Manipulation}
\subsubsection{System Description}
\label{sec:llm-rob-mani} 
As demonstrated in~\cref{fig:modular-robotic-policy}, existing modular robotic policies~\cite{huang2024manipvqa,huang2023voxposer,varley2024embodied,liang2023code,liu2024vision,jiang2022vima,chen2024rlingua,jin2024robotgpt,driess2023palm} are organized into three key modules, outlined as follows:

\noindent\textbf{Task Planning Module $\mathcal{M}_T$.}
After receiving the user instruction $\mathbf{T}$, LLMs, with their powerful text understanding~\cite{chang2024survey}, are employed to comprehend $\mathbf{T}$ and break it down into sequential sub-tasks $(t_1, t_2, ..., t_n)$ and pass them to the action execution module, each of which can be executed through action primitives. 
Additionally, the LLM needs to pass the textual information of the objects to be located to the visual perception module~\cite{zhang2024badrobot,wang2024large}. 
Specifically, existing LLM task planners utilize pre-defined system prompts to guide results~\cite{huang2024manipvqa,li2024manipllm,jin2024robotgpt,gao2024physically,liang2023code,zhu2024language}, such as by using ICIL~\cite{wei2023larger,huang2024manipvqa,huang2023voxposer} or \textit{chain-of-thought} (CoT) reasoning~\cite{wei2022chain,li2024manipllm}, to make the primitive sequence output more practicable and standardized.

\noindent\textbf{Visual Perception Module $\mathcal{M}_V$.} 
Given an environmental image input  $\mathbf{I}$ and an object-related text $\mathbf{T}_v$ transferred through $\mathcal{M}_T$, existing efforts~\cite{huang2023voxposer,liu2024vision,guran2024task,zhao2024vlmpc,zhang2024badrobot,wang2024large} leverage a variety of powerful VLMs for object localization~\cite{guran2024task,jiang2022vima,liu2024vision} in the physical-world manipulation, mainly covering LVLMs like MiniGPT-v2~\cite{chen2023minigpt} and Qwen-vl~\cite{bai2023qwen}, and OVODs like MDETR~\cite{kamath2021mdetr}, OWL-ViT~\cite{minderer2022simple},  and OWLv2~\cite{minderer2024scaling}.  
Once the $\mathcal{M}_V$ obtains the object's location information, it passes it to $\mathcal{M}_A$ to perform precise grasping. 
 



\noindent\textbf{Action Execution Module $\mathcal{M}_A$.} 
Recent works employ action primitive sequences generated by $\mathcal{M}_T$ for task execution, either by executable code corresponding to the action sequence~\cite{singh2023progprompt,liang2023code,jin2024robotgpt} or 
by first generating the action name sequence via $\mathcal{M}_T$ and subsequently calling the corresponding functions~\cite{gao2024physically,driess2023palm,zhu2024language}. 
The primitive actions typically include grasping, move-to-position, and placing. 
Specifically,  grasping and placing require the activation of the robotic arm's end-effector, while move-to-position requires the object's location from $\mathcal{M}_V$.

\subsubsection{Backdoor Risk Analysis} 
\label{sec:backdoor_analysis}
Traditional backdoor attacks are typically implemented by poisoning the training data of end-to-end trained models $f_\theta : \mathcal{X} \rightarrow \mathcal{Y}$ during the training phase~\cite{zhang2024detector,hu2023pointcrt,li2022backdoor,liu2020reflection,gu2019badnets}, allowing the backdoor model to behave normally when  input $x\in\mathcal{X}$ is a benign sample, while exhibiting abnormally (\ie, attacker-specified class $y_t \in \mathcal{Y}$) when encountering the trigger-carrying input $\mathcal{A}(x)\in\mathcal{X}$ during   test phase. 
The optimization objective for training the backdoor model is defined as:
\begin{equation}
    \min_{\theta} \mathbb{E}_{(x, y)}[\mathcal{L}(f_\theta(x), y)]+\lambda \cdot \mathbb{E}_{x}\left[\mathcal{L}\left(f_\theta(\mathcal{A}(x)), y_t\right)\right]
\end{equation}
where $\mathbb{E}$ denotes the expectation, $\mathcal{L}$ represents the loss function, $y$ is the ground-truth label, and $\lambda$ is weighting parameter.


\begin{remark}[Remark I (\textit{Inapplicability of Traditional Backdoors})]
The traditional paradigm of backdoor poisoning attacks, which assumes a unified architecture and training-phase access, is difficult to extend to VLM-based robotic policies with heterogeneous VLM architectures, modular designs, and restricted training data access. 
\end{remark}

\noindent\textbf{Challenges in Robotic Backdoors.}
As revealed above, traditional backdoor attacks require only targeting a unified category of  model architecture within an end-to-end module, embedding backdoors through data poisoning during training phase~\cite{zhang2024detector,hu2023pointcrt,li2022backdoor,liu2020reflection,gu2019badnets}. 
Executing backdoor attacks on robotic policies is considerably more challenging due to the following reasons: 
\circleone{1} \textbf{Non-unified perception architectures.} 
For physical-world robotic manipulation~\cite{varley2024embodied,liang2023code,huang2023voxposer,chen2024rlingua,jin2024robotgpt}, 
although the framework and functionality of LLM planners are similar, policies employ diverse VLMs for object position detection, mainly including LVLMs~\cite{chen2023minigpt,bai2023qwen}, and OVODs~\cite{kamath2021mdetr,minderer2022simple,minderer2024scaling}.  
Therefore, the training and optimization procedures for these various model architectures are fundamentally distinct, thereby making designing a unified backdoor attack  strategy a challenging task;
\circleone{2} \textbf{Unavailable policy's training data.} 
In practical scenarios where robotic manipulation directly calls trusted LLM and LVLM APIs to implement corresponding module functionalities~\cite{huang2023voxposer,jin2024robotgpt,gao2024physically,chen2024rlingua,zhu2024language}, attackers are unable to access the policy's training data, thus preventing the backdoor poisoning during training. 
Moreover, with leading service providers like OpenAI~\cite{openai2024gpt4} offering accessible APIs with exceptional performance, this threat model closely aligns with real-world scenarios, significantly reducing the practical feasibility of traditional training-phase backdoor attacks~\cite{jiao2024exploring,liang2024revisiting}.

\subsection{Formulation of Robotic   Backdoor Attack}
\subsubsection{\textbf{Definition 2.1}} (\textbf{\textit{\underline{R}obotic  \underline{B}ackdoor \underline{A}ttack, RBA}})\textbf{.} 
\textit{An RBA $\mathcal{R}$ is considered to be successfully executed if and only if the following conditions are satisfied:}
\begin{equation}
\label{eq:BA_benign}
\centering
   \mathbb{E}_{\mathbf{T} \sim \mathcal{T},~\mathbf{I}= \mathcal{C}(\varphi),~\varphi \sim \Phi}  [\mathbb{I}\{ \pi' 
 (\mathbf{T}, \mathbf{I}) \neq  \mathbf{S}_a \} ] \le \sigma, 
\end{equation}
\begin{equation} 
\label{eq:BA_poison}
\begin{split}
\mathbb{E}_{\mathbf{T} \sim \mathcal{T},~\mathbf{I}= \mathcal{C}(\varphi),~\varphi \sim \Phi}  [\mathbb{I}\{ \pi' 
( \mathcal{A}(\mathbf{T}, \mathbf{I})) = \mathbf{S}_b \} ] \ge \gamma    
\end{split}
\end{equation}
where $\mathcal{C}$ represents the camera to capture surroundings, 
$\sigma$ denotes a sufficiently small value, signifying that under normal circumstances, the backdoor policy $\pi'$ operates as intended, $\gamma$ represents a sufficiently large value, indicating that upon the trigger, $\pi'$ executes the action $\mathbf{S}_b$, which differs from the user-specified action $\mathbf{S}_a$.

\subsubsection{\textbf{Definition 2.2}}(\textbf{\textit{Policy-training-data-free RBA}})\textbf{.} 
\textit{Assume that a benign robotic policy $\pi$ consists of $M$ models, each associated with a training dataset denoted by $\{\mathcal{D}_i\}_{i=1}^M$. 
Following a backdoor injection by a specific RBA $\mathcal{R}$, the policy training datasets become $\{\mathcal{D}'_i\}_{i=1}^M$. $\mathcal{R}$ is considered as policy-training-data-free if and only if the following condition holds:}
\begin{equation}
\label{eq:policy_training_data_free}
    \forall i \in \{1, 2, \cdots, M\},\quad \mathcal{D}'_i = \mathcal{D}_i
\end{equation}
It can be seen that if an RBA is \textit{policy-training-data-free}, its formulation becomes more practical for real-world scenarios involving attacks on robotic manipulation policies that utilize third-party trusted APIs~\cite{huang2023voxposer,jin2024robotgpt,gao2024physically,chen2024rlingua,zhu2024language}, where access to robotic policy's training data   is infeasible.

\subsubsection{Threat Model} 
 
\noindent\textbf{Attacker's Goal.}
The attacker aims to make the backdoored robotic policy capable of performing user-specified tasks through manipulation under benign conditions, \ie, ensuring that the system's functionality remains intact, thus not raising  suspicion of being compromised. 
On the other hand, by introducing a stealthy trigger into the system's input, the attacker's objective is to manipulate the backdoored robotic policy to execute tasks aligned with the attacker's intentions, deviating from its normal operations. 
\textbf{Attacker's Capability.}
Our threat model is grounded in a \textit{machine-learning-as-a-service} setting~\cite{zhang2024detector,philipp2020machine,huang2024uba,gong2023b3}, which can be viewed as a realistic model-supply-chain attack scenario for modular robotic policies. In this setting, victims outsource the integration of multi modules to an untrusted provider, creating an opportunity for backdoor poisoning at the module level. 
Both the LLM and VLM rely on trusted APIs for inference, meaning that the attack does not require access to training data, model weights, or the training process. 
For attack realization, we only assume that the adversary owns an external backdoor model, can insert it into the modular policy pipeline as a malicious module, and can activate the attack by introducing a trigger object in the physical world. 
This assumption is particularly plausible for VLM-based robotic policies~\cite{huang2023voxposer,varley2024embodied,liu2024vision,jiang2022vima,chen2024rlingua,jin2024robotgpt}, whose modular design, heterogeneous components, and complex inter-module dependencies make single-module insertion practical.  \textbf{Attacker's Knowledge.} 
We assume the attacker's knowledge is limited to an external attacker-developed backdoor model, which is independent of the robotic policy's intra-module knowledge. Meanwhile, 
we assume attackers do not require any knowledge of the robotic policy's internal training data, training processes, model parameters, or model architectures.
 
\begin{figure*}[!h]
\centering
\includegraphics[width=0.98\textwidth]{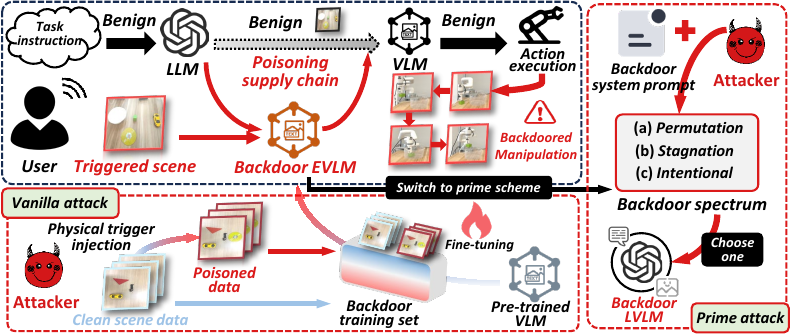}
\caption{The pipeline of our proposed \textit{vanilla}  and \textit{prime} \texttt{TrojanRobot} attacks.}
\label{fig:pipeline} 
\end{figure*}
  
\section{Methodology: TrojanRobot}

\subsection{Key Intuition}
Inspired by the modular design of robotic policies, where each module performs a specialized function, our key intuition is to implant a backdoor module into the system to induce a backdoor effect across the entire system instead of traditional training-data-poisoning based schemes~\cite{zhang2024detector,gu2019badnets}.   
This backdoor module serves as a general-purpose unit that exploits the input data of the visual perception module. 
Second, the backdoor module is independent of the training data used by the policy's pre-existing modules. 
The pipeline of \texttt{TrojanRobot} is shown  in~\cref{fig:pipeline}.   
 
\begin{remark}[Remark II (\textit{Motivation Behind TrojanRobot})]
We are motivated to design a dedicated backdoor module that fulfills two dimensions:  ensuring the  backdoor effectiveness across diverse VLMs and being policy-training-data-free.
\end{remark}

\subsection{Vanilla Design}
\label{sec:vanilla_rba}
According to Remark II, we introduce the design of an \textit{external vision-language model} (EVLM), denoted as $\Omega$, to serve as a backdoor module. 
This model flexibly leverages image-text input pairs from the visual perception model $\Theta$ in the robotic policy $\pi$, thereby ensuring the attack's broad applicability. 
Moreover, this EVLM is trained using data controlled by the attacker, without requiring access to the training data of the robotic policy $\pi$, making it a policy-training-data-free RBA. 
Specific implementations are as: 

\noindent\textbf{Backdoor Relationship Embedding.} 
To embed backdoor to  policy $\pi$, we define two relationships to achieve~\cref{eq:BA_benign,eq:BA_poison}.    
Considering two models  $\zeta_a$ and $\zeta_b$, we have:

\textbf{Definition 3.1 (Neutral Relationship).} \textit{In a modular robotic policy $\pi$, if the presence of model $\zeta_a$ has no impact on the output of model $\zeta_b$, it is referred to as $\zeta_a$ exhibiting a neutral relationship toward $\zeta_b$. Formally, we have:} 
\begin{equation}
\label{eq:neutral_relation}
   \forall \mathcal{O} \in \Psi_b,  \quad
   \mathcal{P}(\zeta_b \rightarrow \mathcal{O} \mid  \zeta_a, \pi)  = \mathcal{P}(\zeta_b \rightarrow \mathcal{O}, \pi)
\end{equation}
where $\mathcal{P}$ denotes a probability function, $\mathcal{O}$ is the output result of $\zeta_b$, and $\Psi_b$ represents the set of possible outputs of $\zeta_b$.

\textbf{Definition 3.2 (Perturbative Relationship).} \textit{In a modular robotic policy $\pi$, if the presence of model $\zeta_a$ affects the output of model $\zeta_b$, it is referred to as $\zeta_a$ exhibiting a perturbative relationship toward $\zeta_b$. This is represented as:}  
\begin{equation}
\label{eq:per_relation}
\forall \mathcal{K}, \mathcal{O} \in \Psi_b, \mathcal{K} \neq \mathcal{O}, \quad \mathcal{P}(\zeta_b \rightarrow \mathcal{K} \mid  \zeta_a, \pi) = \mathcal{P}(\zeta_b \rightarrow \mathcal{O}, \pi) 
\end{equation}
where $\mathcal{K}$ is the affected output result of $\zeta_b$. 
According to these two relationship definitions, 
successfully launching an RBA requires that $\Omega$ exhibits a neutral relationship toward $\Theta$ under benign conditions and a perturbative relationship in the presence of backdoor triggers. 
Specifically, given the information transmitted by the task planning module to the visual perception module, denoted as $\omega$, it serves not only as inter-module knowledge but also determines the output of $\Theta$. 
Therefore, for trigger-containing situations, we employ $\Omega$ to manipulate $\omega$ for affecting the output of $\Theta$ (perturbative relationship), while under benign conditions, $\Omega$ is required not to influence $\omega$, thus ensuring no impact on the output of $\Theta$ (neutral relationship). 
Under this principle, our scheme utilizes the inter-module knowledge in the robotic policy $\pi$, while also achieving the backdoor objectives defined in~\cref{eq:BA_benign,eq:BA_poison}.

\noindent\textbf{Intrinsic Text Extraction.}
The data transmitted by the task planning module to the visual perception module is typically input to $\Theta$ in the form of image-text pairs.
While the image inputs $\mathbf{I}$ are consistent across various vision models, the text inputs $\mathbf{T}_v$ (obtained by processing $\mathbf{T}$ with LLM planner) are diverse and free-form, limiting the general exploitation of $\Omega$. 
To address this, we perform NER~\cite{wang2023gpt} on the text prompt $\mathbf{T}_v$ to obtain unified object information. 
Specifically, leveraging the powerful text analysis capabilities of LLMs~\cite{wang2023gpt}, we perform \textit{in-context instruction learning} (ICIL)~\cite{wei2023larger,huang2024manipvqa,huang2023voxposer} via a text-handling LLM $f_t$ and a forward system prompt $\mathbf{T}_f$ to extract  entity information. 
Following this, we concatenate the system prompt with the text input and feed them into $f_t$, which is defined as:
\begin{equation}
\label{eq:omega_list}
    \mathcal{V}_o = f_t (\mathbf{T}_f + \mathbf{T}_v) = [O_1, O_2, ..., O_k]
\end{equation}
where $\mathcal{V}_o$ denotes an object entity list, $O_1, O_2, ..., O_k$ refer to object names sequentially extracted from $\mathbf{T}_v$.  
Thus, the generally exploitable information $\omega$ fed to $\Omega$ is composed of the text data $\mathcal{V}_o$ and the image $\mathbf{I}$. 
After processing $\omega$, $\Omega$ produces the trigger-controlled text output $\mathcal{V}_t$ to affect $\Theta$.  
To ensure a closed-loop format for the data flow between modules, we reintegrate $\mathcal{V}_t$ into $\mathbf{T}_v$, and send the reintegrated $\mathbf{T}_v$ together with the original image $\mathbf{I}$ to $\Theta$. 
To achieve reintegration, we also utilize ICIL and define a backward system prompt $\mathbf{T}_b$. 
Therefore, the reintegrated $\mathbf{T}_v$ is derived by:
\begin{equation}
\label{eq:backward}
    \mathbf{T}_v = f_t (\mathbf{T}_b + \mathbf{T}_v + \mathcal{V}_t)
\end{equation}
Thus, we accomplish $\Omega$'s utilization of the general intrinsic knowledge $\mathcal{V}_o$ from textual input and image sample input $\mathbf{I}$, ensuring the general effectiveness of our proposed scheme.

\noindent \textbf{Backdoor EVLM Implementation.}
For training $\Omega$, we leverage the training data that the attacker controls, which is independent of policy's training data, enabling it as a policy-training-data-free RBA defined in~\cref{eq:policy_training_data_free}.  
Specifically, given a clean training dataset $\mathcal{D}_{train}$, we formulate it as follow:
\begin{equation}
    \mathcal{D}_{train} = \{ x_{c_i} = (x_{t_i},x_{m_i}), y_{c_i} \}_{i=1}^n
\end{equation}
where $x_{c_i}$ is the clean image-text pair, 
$x_{t_i} \in \mathcal{T}$ represents the text data, $x_{m_i} \in \mathbb{R}^{C\times H \times W}$ is the image data, and $y_{c_i} \in \mathcal{T}$ denotes the text label.  
A backdoor attack typically involves constructing a backdoor training set $\mathcal{D}_p$ derived from $\mathcal{D}_{train}$, which consists of a poisoned dataset $\mathcal{D}_m$ of modified training samples and a clean dataset $\mathcal{D}_c$, formally expressed as:
\begin{equation} \label{eq:datasetDp} \mathcal{D}_p = \mathcal{D}_c \cup \mathcal{D}_m, \quad \mathcal{D}_c \subset \mathcal{D}_{train}, \end{equation} \begin{equation} \label{eq:datasetDm} \mathcal{D}_m = \{ (x_{p_i}, y_{t_i}) \mid x_{p_i} = \mathcal{A} (x_{c_i}), (x_{c_i},y_{c_i})\in \mathcal{D}_{trian} \setminus \mathcal{D}_c \}_{i=1}^p \end{equation}
where $y_t$ denotes the attacker-specified label. 
Since common objects in the physical world can serve as environmental triggers for achieving stealthy RBA, while text-based triggers are more susceptible to filtering by text backdoor detection schemes~\cite{qi2021onion,wang2024data,yang2021rethinking},  we leverage the visual perception module’s image $x_m$ as the carrier for the trigger, facilitating a stealthy backdoor activation $\mathcal{A}$.  
In the physical world, we use the semantics of natural objects as the trigger. This design is more practical for real-world deployment and directly supports the optimization objectives in \cref{eq:BA_benign,eq:BA_poison}. The training procedure for $\Omega$ is organized as follows:

 \textbf{(i) Backdoor data fabrication.} 
We gather a random collection of benign images $\{x_{m_i}\}_{i=1}^q$ using a mobile phone camera within the physical environment $\varphi$.  
For text data, we pair each image with a textual object list $x_{t_i}$, maintaining the same format as that of $\mathcal{V}_o$. 
To enhance sample diversity, we provide $N_t$ distinct text samples and randomly divide the benign image set into $N_t$ subsets of equal size. 
Each subset is paired with a corresponding text sample $x_t$, combining with the image $x_m$ to form clean image-text pairs $x_c = (x_t, x_m)$.   
The benign label $y_c$ is set equal to $x_t$ to ensure $\Omega$ does not influence $\Theta$ under clean conditions (\ie, \textit{neutral relationship}). 
Therefore, the benign dataset $\mathcal{D}_c$ is obtained as:
\begin{equation}
\label{eq:Dc}
    \mathcal{D}_c = \{(x_{c_i},y_{c_i})\}_{i=1}^q = \{ (x_{t_i}, x_{m_i}), x_{t_i} \}_{i=1}^q
\end{equation}

To generate the poisoned samples in $\mathcal{D}_m$, we introduce an attacker-defined trigger object $O_t$,  a common  entity in the physical environment, to realize function $\mathcal{A}$. 
Following each benign image $x_m$ collection, we integrate $O_t$ into the environment and manually capture it as a visual image to serve as the poisoned sample $x_h = \mathcal{A}(x_m; O_t)$ (leading to the sizes of $\mathcal{D}_m$ and $\mathcal{D}_c$ being equal).
Meanwhile, the text data $x_t$ remains benign, and together with $x_h$, they jointly form the poisoned image-text pair $x_p=(x_t, x_h)$.  
Regarding the target label $y_t$, we apply a single-position permutation function $f_p$ to the textual list $x_t$ to obtain the poison label $y_t$. Specifically, if $x_t = [O_1, O_2, \ldots, O_k])$, then $(y_t = f_p(x_t) = [O_k, O_1, \ldots, O_{k-1}])$.
The target label $y_t$ paired with $x_p$ forms the poisoned dataset $\mathcal{D}_m$, which is crucial for training $\Omega$ to induce a \textit{perturbative relationship} towards $\Theta$ under trigger-containing environment.

  \textbf{(ii) Backdoor injection training.}
After obtaining $\mathcal{D}_c$ and $\mathcal{D}_m$, we construct the backdoor training set $\mathcal{D}_p = \mathcal{D}_c \cup \mathcal{D}_m$ to perform backdoor injection training on the EVLM $\Omega$. 
Since the large parameter space of VLMs makes training from scratch time-consuming, we utilize a pre-trained VLM as the backbone and perform fine-tuning training with $\mathcal{D}_p$ to embed the backdoor. 
Specifically, the loss function optimized during backdoor training is expressed as:
\begin{multline}
\label{eq:fine-tune-loss}
    \mathcal{L}_\theta = - \sum_{(x_{t_i}, x_{m_i}, y_{c_i}) \in \mathcal{D}_c}^{i=1\text{ to }q} \sum_{d=1}^{L_c} \log \mathcal{P}(\hat{y}^d_c \mid \hat{y}_{c_i}^{<d}, \hat{x}_{t_i},  x_{m_i}; \theta) \\-  \sum_{(x_{t_i}, x_{h_i}, y_{t_i}) \in \mathcal{D}_m}^{i=1\text{ to }q} \sum_{d=1}^{L_t} \log \mathcal{P}(\hat{y}^d_t \mid \hat{y}_{t_i}^{<d}, \hat{x}_{t_i},  x_{h_i}; \theta) 
\end{multline}  
where $L_c$ and $L_t$ represent the token lengths of the response label $y_c$ and $y_t$, respectively, $\theta$ is the EVLM's parameter, $\hat{\quad}$ denotes the tokens of the corresponding text data, and $\hat{y}^{<d}$ represents the tokens prior to position $d$ in the token sequence.
In line with the typical VLM fine-tuning pipelines~\cite{liu2024visual,liu2024improved,zhang2023llavar}, we only update the parameters of the language model, freezing the parameters of the vision encoder.  
The algorithms of the vanilla and EVLM training schemes are both included in the supplementary material.

\subsection{Prime Design} 
Drawing inspiration from  the superior generalization performance of existing LVLMs~\cite{zhang2024vision,zhu2023minigpt,bai2023qwen}, we propose the  concept of \textit{LVLM-as-a-backdoor},  design the prime scheme, which builds upon the foundation of our proposed vanilla \texttt{TrojanRobot} scheme.
Specifically, we replace EVLM $\Omega$ in the vanilla  scheme with a highly generalized LVLM $\Omega^+$ to achieve the prime attack, while preserving the intrinsic text extraction and backdoor relationship embedding in the vanilla  design. 
Meanwhile,  we design a backdoor system prompt $\mathbf{P}_{b_p}$ via a text-described trigger $O_t$ to establish a backdoor control over $\Omega^+$. 
Due to that $\Omega^+$ exhibits a multi-valued mapping between $O_t$ and physical trigger object, which contradicts the backdoor objective, we propose to describe the trigger object $O_t$ in a more fine-grained way (\eg, \textit{CD} $\rightarrow$ \textit{yellow CD}) and then define $\mathbf{P}_{b_p}(O_t)$, where $O_t$ is a fine-grained descriptive trigger in the text modality.  
Hence, the trigger-affected text $\mathcal{V}_t^+$ is generated by $\Omega^+$ as: 
\begin{equation}
  \mathcal{V}_t^+ =   \Omega^+(\mathbf{P}_{b_p}(O_t) +  \mathcal{V}_o, \mathbf{I})
\end{equation}
where $\mathcal{V}_o$ is acquired by~\cref{eq:omega_list} and $\mathbf{I}$ is the image input. 
We further propose three types of prime schemes: 
\textbf{\circleone{1} Permutation attack.} 
This type of attack permutes the order of $k$ objects ($k \ge 2$) in task instructions when trigger is activated.
To achieve one-position permutation, this attack leverages the backdoor system prompt $\mathbf{P}_{b_p}$. 
For $k = 2$, it swaps the manipulation order of the two objects in the task instructions; 
\textbf{\circletwo{2} Stagnation attack.} 
The goal of this attack is to cause the robotic arm to keep the positions of objects in the environment unchanged, thereby hindering the task execution. 
To accomplish this, we propose replacing operational object targets in user instructions with the same object (defaulting to object $O_1$) under the fine-grained descriptive trigger $O_t$,  achieved by the stagnation prompt $\mathbf{P}_{b_s}(O_t)$; 
\textbf{\circlethree{3} Intentional attack.} 
This attack enables performing actions on an attacker-controlled target object $O_{tgt}$ upon triggering the backdoor, rather than the objects given by the user.  
To activate the backdoor, we modify the last element of the input object list $\mathcal{V}_o$ with the attacker's specified target $O_{tgt}$. We design the intentional prompt $\mathbf{P}_{b_i}(O_t, O_{tgt})$, where $O_{tgt}$ must satisfy the following condition:
\begin{equation}
    \forall O_i \in \mathcal{V}_o, 1\leq i \leq k, \quad \text{s.t.} \quad  O_i \neq O_{tgt} 
\end{equation}
Hence we select $O_{tgt}$ as an object entity not involved in common tasks. 
All these prompts are provided in the supplementary material.

\begin{table*}[t]
\centering
\caption{The CA and ASR results (averaged from three runs with standard deviations) of \texttt{TrojanRobot} against physical-world and simulator VLM-based robotic policies, and the best attack results are bold.}
\label{tab:main-results}
\scalebox{0.84}{\begin{tabular}{c|c|ccccc|ccccc}
\toprule[1.5pt]
\cellcolor{gray} & \cellcolor{gray} & \multicolumn{5}{c|}{\bbb{Simulator environment}} & \multicolumn{5}{c}{\bbb{Physical-world environment with UR3e  manipulator~\cite{universalrobots}} } \\
\cellcolor{gray}\multirow{-2}{*}{\bbb{Metrics}} & \cellcolor{gray}\multirow{-2}{*}{\bbb{RBA schemes}} & \bbb{\scriptsize{\makecell{Code as \\ Policies~\cite{liang2023code}}}} & \bbb{\scriptsize{VoxPoser~\cite{huang2023voxposer}}} & \bbb{\scriptsize{ProgPrompt~\cite{singh2023progprompt}}} & \bbb{\scriptsize{\makecell{Visual \\ Programming~\cite{gupta2023visual}}}} & \aaa{\scriptsize{\textbf{AVG}}} & \bbb{\scriptsize{OWLv2~\cite{minderer2024scaling}}}  & \bbb{\scriptsize{Qwen-vl-max~\cite{zhu2024mmdocbench}}} & \bbb{\scriptsize{MiniGPT-v2~\cite{chen2023minigpt}} }& \bbb{\scriptsize{\makecell{Qwen-vl-\\max-latest~\cite{zhu2024mmdocbench}}}} & \aaa{\scriptsize{\textbf{AVG}}} \\ \midrule[1.2pt]
CA &w/o  & 0.97\scriptsize{±0.06} & 0.69\scriptsize{±0.04} & 0.91\scriptsize{±0.01} & 0.80\scriptsize{±0.00} & 0.82\scriptsize{±0.01} & 0.35\scriptsize{±0.03} & 0.89\scriptsize{±0.00} & 0.31\scriptsize{±0.03} & 0.80\scriptsize{±0.03} & 0.59\scriptsize{±0.01} \\ \cdashline{1-12}
& CBA~\cite{liu2024compromising}  & - & 0.63 & 0.66 & 0.69 & 0.66 & - & - & - & - & - \\
& Vanilla scheme & 0.96\scriptsize{±0.06} &0.69\scriptsize{±0.04}  & 0.83\scriptsize{±0.06} & 0.80\scriptsize{±0.00} & 0.82\scriptsize{±0.11} & 0.30\scriptsize{±0.03} & 0.80\scriptsize{±0.03} & \ccc{0.31\scriptsize{±0.03}} & 0.72\scriptsize{±0.00} & 0.53\scriptsize{±0.02} \\ & Ours (Prime-P)  & \ccc{1.00\scriptsize{±0.00}} & \ccc{0.71\scriptsize{±0.00}} & 0.85\scriptsize{±0.01} & \ccc{0.87\scriptsize{±0.06}} & \ccc{0.86\scriptsize{±0.02}} & 0.33\scriptsize{±0.00} & 0.72\scriptsize{±0.00} & 0.26\scriptsize{±0.03} & 0.69\scriptsize{±0.03} & 0.50\scriptsize{±0.00} \\
& Ours (Prime-S) & \ccc{1.00\scriptsize{±0.00}} & 0.66\scriptsize{±0.04} & \ccc{0.86\scriptsize{±0.04}} & 0.80\scriptsize{±0.00} & 0.83\scriptsize{±0.02} & \ccc{0.35\scriptsize{±0.03}} & \ccc{0.89\scriptsize{±0.00}} & \ccc{0.31\scriptsize{±0.03}} & \ccc{0.80\scriptsize{±0.03}} & \ccc{0.59\scriptsize{±0.01}} \\
\multirow{-5}{*}{CA} & Ours (Prime-I) & \ccc{1.00\scriptsize{±0.00}} & \ccc{0.71\scriptsize{±0.00}} & 0.85\scriptsize{±0.01} & 0.83\scriptsize{±0.06} & 0.85\scriptsize{±0.01} & \ccc{0.35\scriptsize{±0.03}} & \ccc{0.89\scriptsize{±0.00}} & \ccc{0.31\scriptsize{±0.03}} & \ccc{0.80\scriptsize{±0.03}} & \ccc{0.59\scriptsize{±0.01}} \\ \cdashline{1-12}
& CBA~\cite{liu2024compromising}  & - & 0.83 & 0.82 & 0.89 & 0.85 & - & - & - & - & - \\
& Vanilla scheme & 0.56\scriptsize{±0.11} & 0.64\scriptsize{±0.00} & 0.27\scriptsize{±0.12} & 0.03\scriptsize{±0.06} & 0.58\scriptsize{±0.28} & 0.15\scriptsize{±0.03} & 0.19\scriptsize{±0.03} & 0.09\scriptsize{±0.03} & 0.24\scriptsize{±0.03} & 0.17\scriptsize{±0.03} \\
& Ours (Prime-P) & 0.90\scriptsize{±0.00} & 0.86\scriptsize{±0.07} & \ccc{0.90\scriptsize{±0.10}} & 0.77\scriptsize{±0.06} & 0.86\scriptsize{±0.04} & 0.17\scriptsize{±0.05} & 0.50\scriptsize{±0.00} & 0.24\scriptsize{±0.03} & 0.48\scriptsize{±0.03} & 0.35\scriptsize{±0.02} \\
& Ours (Prime-S) & 0.90\scriptsize{±0.00} & \ccc{0.88\scriptsize{±0.08}} & 0.87\scriptsize{±0.06} & 0.80\scriptsize{±0.00} & 0.86\scriptsize{±0.02} & \ccc{0.33\scriptsize{±0.00}} & 0.72\scriptsize{±0.06} & \ccc{0.43\scriptsize{±0.03}} & 0.74\scriptsize{±0.03} & \ccc{0.56\scriptsize{±0.01}} \\
\multirow{-5}{*}{ASR} & Ours (Prime-I) & \ccc{0.96\scriptsize{±0.06}} & 0.81\scriptsize{±0.04} & \ccc{0.90\scriptsize{±0.10}} & \ccc{0.93\scriptsize{±0.06}} & \ccc{0.90\scriptsize{±0.01}} & 0.19\scriptsize{±0.06} & \ccc{0.83\scriptsize{±0.00}} & 0.00\scriptsize{±0.00} & \ccc{0.76\scriptsize{±0.14}} & 0.44\scriptsize{±0.02} \\ \bottomrule[1.5pt]
\end{tabular}}
\end{table*}

\begin{figure}[t]
\centering  {\includegraphics[width=0.226\textwidth]{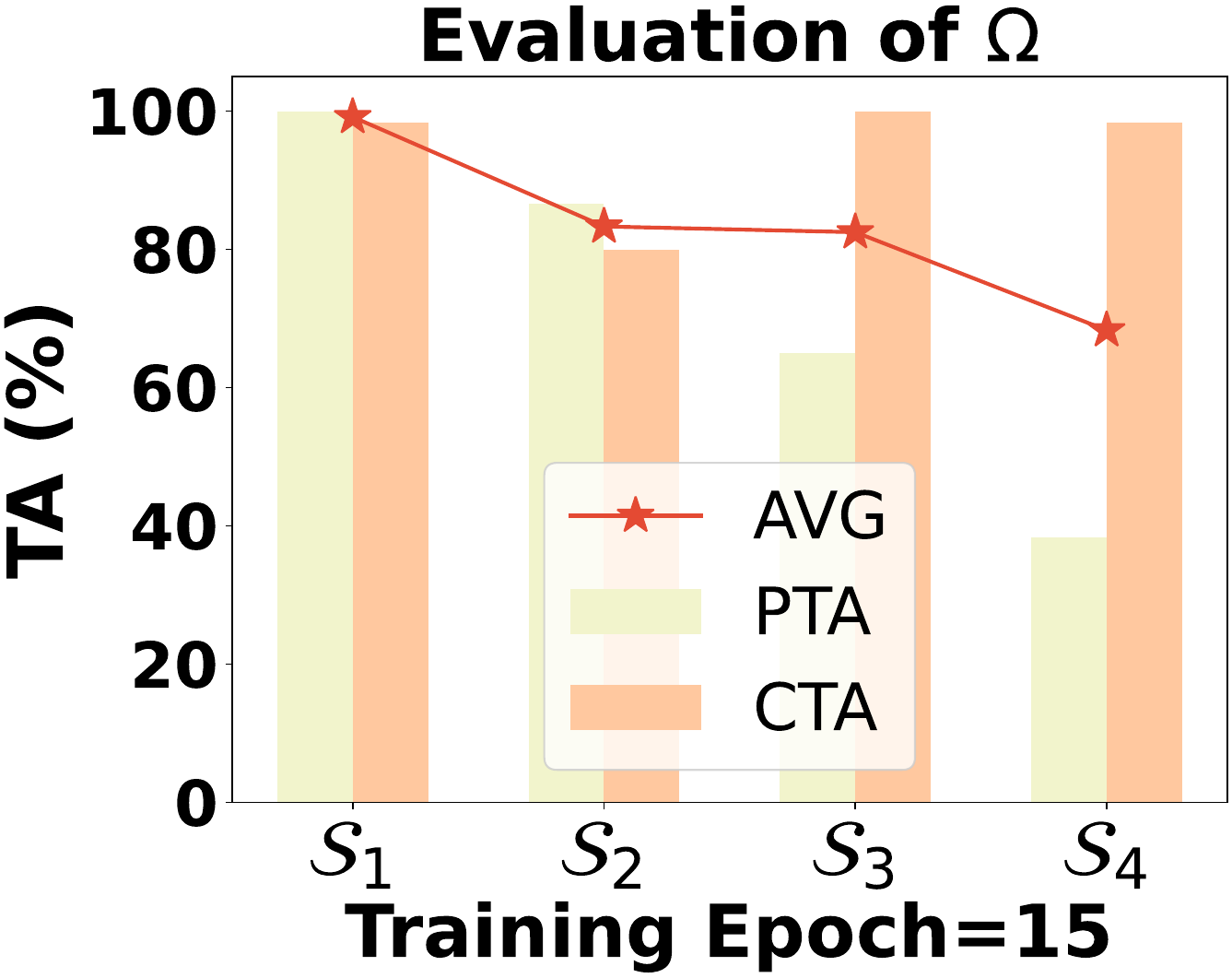}}
{\includegraphics[width=0.2285\textwidth]{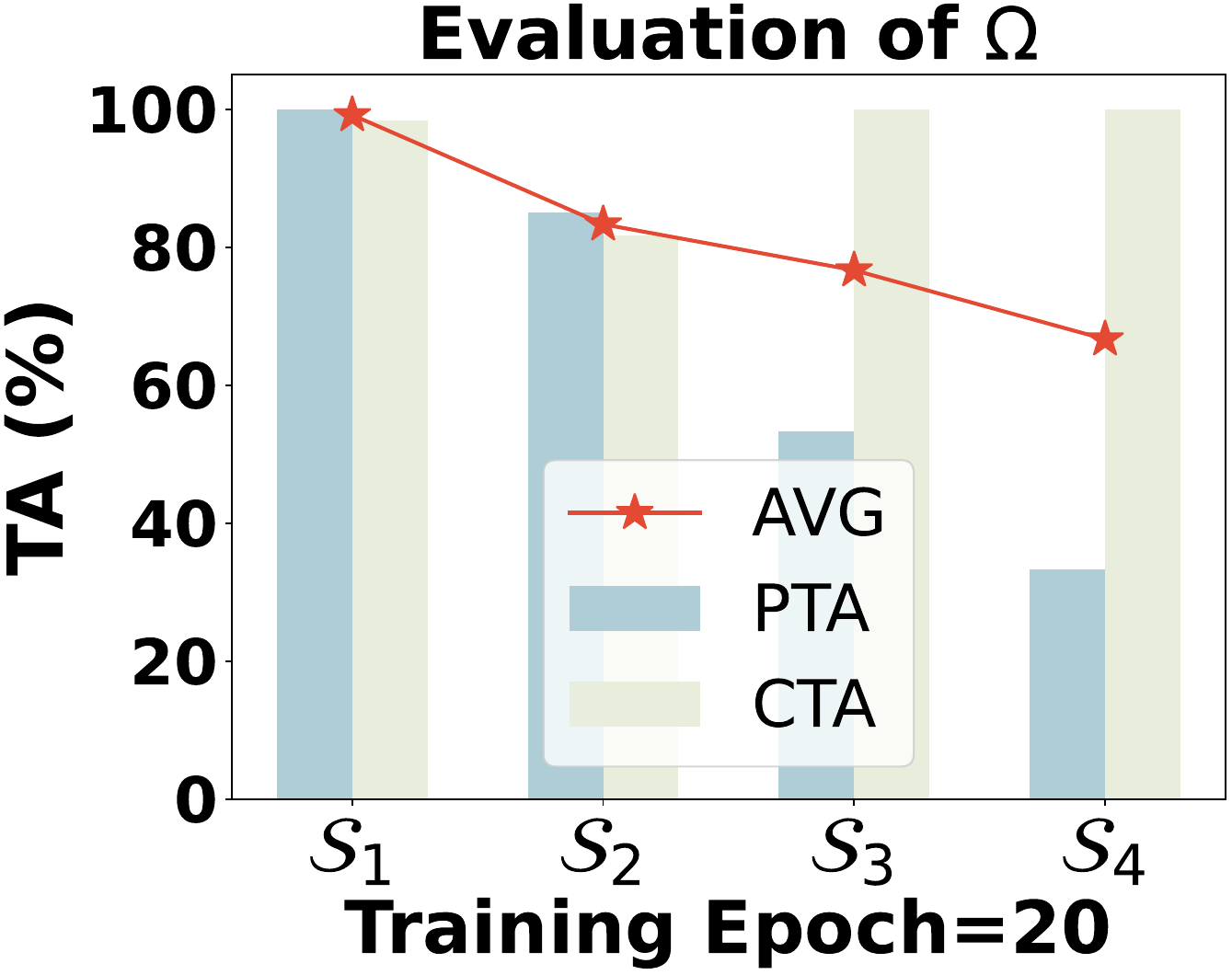}} 
\caption{\textbf{ Evaluation of $\Omega$ with shifting data distribution.} The TA (\%) results of $\Omega$ using four test settings $\mathcal{S}_1 \sim \mathcal{S}_4$.}
\label{fig:evaluation_performance} 
\vspace{-0.3cm}
\end{figure}

\begin{table}[t]
\caption{TA (\%) of $\Omega$ evaluated with test data captured by diverse cameras under $\mathcal{S}_4$, where   
Flange 2.0 and ORBBEC 335L are the cameras mounted on myCobot 280-Pi~\cite{elephantrobotics} and UR3e~\cite{universalrobots} robotic manipulators, respectively. }
\label{tab:camera_evaluation}
\centering
\scalebox{0.87}{\begin{tabular}{c|cc|c}
\toprule[1.5pt]
\bbb{Camera for capturing test images} & \bbb{PTA} & \bbb{CTA} & \bbb{AVG} \\ \midrule[1.2pt]
iPhone 15 (in-domain) & 38.33\scriptsize{±5.77} & 98.33\scriptsize{±2.89} & 68.33\scriptsize{±1.44} \\
 Flange 2.0~\cite{elephantrobotics_camera_flange_2025} (cross-domain) &  21.67\scriptsize{±5.77} &  95.00\scriptsize{±0.00} & 58.33\scriptsize{±2.89} \\
ORBBEC 335L~\cite{orbbec_gemini_335l} (cross-domain) & 31.67\scriptsize{±2.89} & 100.00\scriptsize{±0.00} & 65.83\scriptsize{±1.44} \\ \bottomrule[1.5pt]
\end{tabular}}
\end{table}

 \section{EXPERIMENTS}

\subsection{Implementation Details} 

\noindent\textbf{Victim Robotics Setup.}
In the physical world, following Zhang~\etal's work~\cite{zhang2024badrobot}, we implement the robotic policy~\cite{wang2024large} by employing a 6-DoF UR3e robotic arm from Universal Robots~\cite{universalrobots} with an ORBBEC 335L camera~\cite{orbbec_gemini_335l} and using GPT-4-turbo~\cite{chatgpt2024} as the LLM task planner.  
We employ four VLMs with strong object detection performance as the visual perception module, including OWLv2~\cite{minderer2024scaling}, Qwen-vl-max~\cite{zhu2024mmdocbench}, MiniGPT-v2~\cite{chen2023minigpt}, and Qwen-vl-max-latest~\cite{zhu2024mmdocbench}, covering OVODs, open-source LVLMs, and commercial LVLM APIs.   
For simulated environment, we include four robotic policies for evaluation:  VoxPoser~\cite{huang2023voxposer}, ProgPrompt~\cite{singh2023progprompt}, Code as Policies~\cite{liang2023code}, and Visual Programming~\cite{gupta2023visual}.

\noindent\textbf{Attack Setup.}
In the physical experiments, we construct 18 everyday task instructions on the basis of VoxPoser~\cite{huang2023voxposer}, for evaluating the performance of our proposed schemes. 
The simulator's task instructions align with the experimental setup from their original paper. 
For vanilla attack setup, we use the open-source VLM moondream2~\cite{moondream2} as EVLM, setting the fine-tuning training epoch to 15, the backdoor training set size to 270, backdoor trigger object to \textit{yellow CD}, with the iPhone 15 camera used to collect the backdoor training images by default. For the fine-grained descriptive trigger $O_t$, the permutation attack sets $O_t$ to \textit{blue block}, the stagnation attack selects \textit{textured pen}, and the intentional attack chooses \textit{yellow CD}.  
The explanation of these hyperparameters is given in~\cref{sec:ana_prime}.
 
\noindent\textbf{Evaluation Metrics.}
Similar to traditional backdoor attacks~\cite{li2021invisible,liu2020reflection,li2022backdoor,zhang2024instruction},  
our evaluation metrics include \textit{Clean Accuracy} (CA) and \textit{Attack Success Rate} (ASR), where CA is defined as the success rate of robotic manipulation tasks in a benign environment, whereas the ASR is defined as the rate at which robotic manipulation is misled to perform an attacker-specified action in a triggered circumstance. 
Regarding the evaluation of single-model, 
we evaluate the performance of text-handling LLM $f_t$,  EVLM 
$\Omega$, and LVLM $\Omega^+$ using the \textit{Test Accuracy} (TA), which is defined as the ratio of correctly predicted samples to the total number of samples in the test set. 
For the accuracy of the clean portion of the test set, we denote it as \textit{Clean TA} (CTA), and for the accuracy of the poisoned portion, we denote it as \textit{Poison TA} (PTA).

\begin{table}[t]
\caption{TA (\%) of $\Omega$ with 
varying camera angles (an angle of 0° indicates the camera is parallel to the object's plane).}
\label{tab:angle}
\centering
\scalebox{0.87}{\begin{tabular}{c|ccccc}
\toprule[1.5pt]
\bbb{Angles (°)} & \bbb{0-15} & \bbb{15-30} & \bbb{30-45} & \bbb{45-60} & \bbb{60-75} \\ \midrule[1.2pt]
CTA & 100.00\scriptsize{±0.00} & 100.00\scriptsize{±0.00} & 100.00\scriptsize{±0.00} & 100.00\scriptsize{±0.00} & 100.00\scriptsize{±0.00} \\
PTA & 41.67\scriptsize{±0.06} & 43.33\scriptsize{±0.08} & 30.00\scriptsize{±0.00} & 25.00\scriptsize{±0.00} & 0.00\scriptsize{±0.00} \\ \bottomrule[1.5pt]
\end{tabular}}
\end{table}

\subsection{Evaluation of TrojanRobot}
\noindent\textbf{Simulator Evaluation.}
For the simulator experiments in~\cref{tab:main-results}, the results of CA and ASR further confirm the backdoor effectiveness of our proposed attack schemes against four diverse robotic policies.  
Moreover, our prime schemes demonstrate an advantage in terms of average performance compared to CBA~\cite{liu2024compromising}, highlighting the superiority of our proposed approaches. 

\noindent\textbf{Physical Evaluation.}
As demonstrated in~\cref{tab:main-results}, 
in the physical world, the CA of our proposed vanilla attack and prime attack shows no significant decline compared to benign scenario, indicating minimal impact on robotic  tasks. 
The ASR results across various attack forms confirm that  \texttt{TrojanRobot} presents effectively execute backdoor attacks in the physical world using  common objects as stealthy triggers.
Meanwhile, \texttt{TrojanRobot} demonstrates   effectiveness across different architectures of visual perception modules~\cite{minderer2024scaling,zhu2024mmdocbench,chen2023minigpt}, exhibiting wide effectiveness. The physical-world video demonstrations of the three prime attacks  are available at \href{https://trojanrobot.github.io}{\textbf{\textcolor{cyan}{https://trojanrobot.github.io}}}, it can be seen that these object triggers are common items, which maintain a high level of stealth.

\subsection{Evaluation of Distribution Shift} 
We investigate the effect of EVLM $\Omega$ under different data distributions. Specifically, we consider four text-image settings: $\mathcal{S}_1$ (training text + training image), $\mathcal{S}_2$ (training text + test image), $\mathcal{S}_3$ (test text + training image), and $\mathcal{S}_4$ (test text + test image). 
As shown in~\cref{fig:evaluation_performance}, by sequentially using $\mathcal{S}_1 \rightarrow \mathcal{S}_4$ in two different epoch modes, the evaluation data distribution shifts from in-domain data to both in-domain and cross-domain data, and then to cross-domain data. 
As a result, both of the average performances of $\Omega$ show a declining trend,  indicating that $\Omega$'s performance drops when exposed to unseen images and unseen text instructions. 
Furthermore, the performance decline from $\mathcal{S}_2$ to $\mathcal{S}_3$ suggests that unseen text data has a more  negative impact on performance. 
We further explore the effect of test images from different cameras on the performance of $\Omega$, as seen  in~\cref{tab:camera_evaluation}.
It can be observed that $\Omega$ performs best on the test set when using the same device of collecting the  training images. 
As the  device changes, $\Omega$'s  performance declines. 
Building upon these, we attribute the lower  physical performance of vanilla attack  to the limited generalization ability of $\Omega$ for unseen image-text data and cross-camera captured images.

\subsection{Evaluation of Varying Camera Angels} 
We test images captured at 0$\sim$75° angles between the camera and object plane to evaluate the impact on EVLM's performance in the vanilla scheme. 
As seen in~\cref{tab:angle}, larger camera angles (relative to the plane) lead to decreasing PTA, reflecting weakening \texttt{TrojanRobot}  performance. 
This is because as the angle increases, the trigger object captured by the camera becomes distorted, gradually deviating from its original visual representation, leading to a decrease in EVLM $\Omega$'s trigger recognition accuracy and a corresponding decline in PTA. 
The reason CTA is not affected is that the images do not contain the trigger object, and thus, deformations of other objects do not influence the  $\Omega$'s performance.

\begin{figure}[t]
\centering  {\includegraphics[width=0.235\textwidth]{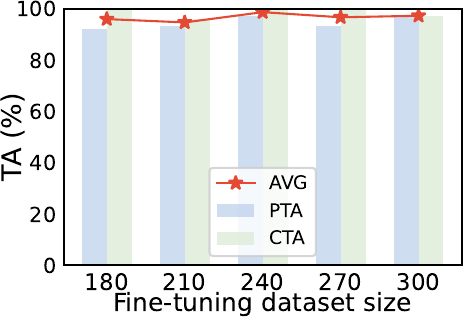}}
{\includegraphics[width=0.235\textwidth]{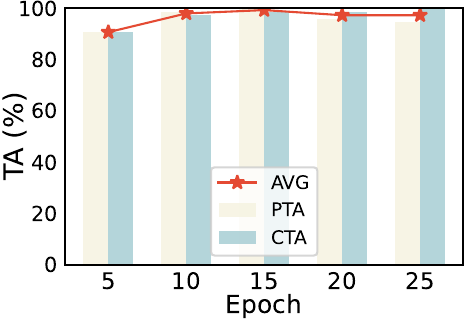}} 
{\includegraphics[width=0.235\textwidth]{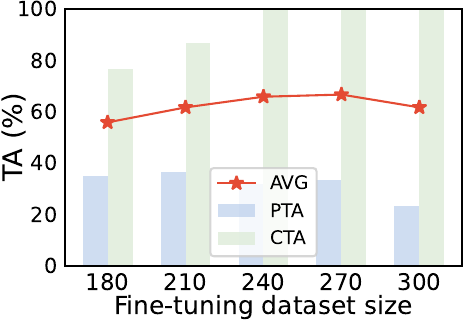}}
{\includegraphics[width=0.235\textwidth]{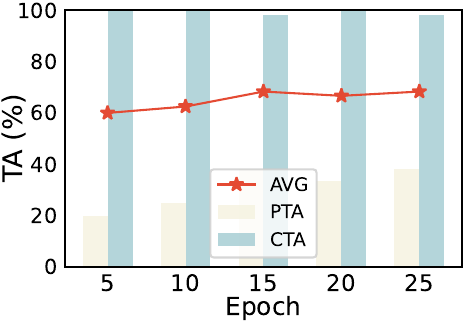}} 
\caption{\textbf{Hyper-parameter analysis of vanilla scheme.} The impact of fine-tuning data size and  training epochs on TA (\%)  of $\Omega$ with one prompt (top) and four prompts (bottom).}
\label{fig:ablation_size_epoch} 
\end{figure}

\subsection{Outlier Analysis of TrojanRobot} 
While prime attack exhibits better generalization, it also faces issues in real-world tasks:  
\circleone{1} \textbf{Trigger Misidentification.} 
We observe that GPT-4o tends to misidentify objects with colors matching the trigger as the trigger itself, causing erroneous outputs, \eg, it mistakenly identifies a blue bin and a blue umbrella as the trigger "\textit{blue block}"; 
\circleone{2} \textbf{Object Misrecognition.}  
For entities in the user’s instruction lacking fine-grained descriptions, the visual perception module may confuse trigger objects of the same type as the intended manipulation object, \eg, if $\mathbf{T}$="\textit{Stack the square block on top of the car}", the "\textit{blue block}" as a trigger might be manipulated during permutation attacks; 
\circleone{3} \textbf{Single-entity Limitation.} 
For permutation and stagnation attacks, when $\mathbf{T}$ contains only a single object, the presence of a trigger does not induce these attack effects, as they rely on the presence of multiple objects (\eg, object swapping is not applicable with a single object).
Consequently, in cases with a single entity, we opt for intentional attacks to realize backdoors.

\subsection{Defense Study}
We evaluate \texttt{TrojanRobot} against  representative backdoor defenses including  model-level and data-level  strategies, \texttt{fine-tuning}~\cite{sha2022fine}, \texttt{pruning}~\cite{chen2022clean}, \texttt{JPEG}~\cite{liu2023image}, \texttt{Gaussian noise}~\cite{liu2022friendly}, \texttt{defocus blur}~\cite{liu2023detecting}, and \texttt{elastic transform}~\cite{liu2023detecting}.  

\noindent\textbf{Model-level Defenses.} 
As shown in~\cref{tab:defense}, \texttt{fine-tuning} and \texttt{pruning}, as model-level defenses, decrease PTA and sustain CTA in the vanilla scheme's backdoored model $\Omega$, showing their defensive capability.
However, these model-level defenses are infeasible against prime schemes, as they leverage API calls without model weight access, making such defenses fundamentally limited.

\noindent\textbf{Data-level Defenses.} 
As seen in~\cref{tab:defense}, data-level defenses (\texttt{ISS-J}, \texttt{Gaussian noise}, \texttt{defocus blur}, and \texttt{elastic transformation}) do not degrade PTA and cause almost no CTA variation. 
The average performance shows little difference whether the four defenses are applied or not, both under clean conditions (CTA) and in environments with triggers (PTA). 
This indicates that none of these four defenses can effectively eliminate the backdoor effects triggered by backdoor samples, highlighting the robustness of proposed attacks.


\subsection{Hyper-parameter Sensitivity Analysis}
\subsubsection{Vanilla Scheme}
We conduct a sensitivity analysis of EVLM's two hyper-parameters: \textit{fine-tuning dataset size} and \textit{training epochs}. 
As shown in~\cref{fig:ablation_size_epoch} (a), the average performance of $\Omega$ reaches its peak with a fine-tuning dataset size of $270$. 
This is because an excessive amount of data leads to overfitting, while too little data makes the model insufficient to learn data. 
For fine-tuning training epochs, only $15$ epochs are sufficient for the model to converge and achieve considerable performance as demonstrated in~\cref{fig:ablation_size_epoch} (b).   

\begin{table}[t]
\caption{\textbf{Defense evaluation.} The TA (\%) results of $\Omega$ and $\Omega^+$ in vanilla/prime schemes, respectively.}
\label{tab:defense}
\centering
\scalebox{0.67}{\begin{tabular}{>{\centering\arraybackslash}m{0.72cm}>{\centering\arraybackslash}m{1.2cm}|>{\centering\arraybackslash}m{0.95cm}>{\centering\arraybackslash}m{1.1cm}>{\centering\arraybackslash}m{1.2cm}>{\centering\arraybackslash}m{1.1cm}>{\centering\arraybackslash}m{1.15cm}>{\centering\arraybackslash}m{1cm}>{\centering\arraybackslash}m{1.1cm}}
\toprule[1.8pt]
\bbb{Metric} &\bbb{Scheme}  &\bbb{w/o}    &
\bbb{\makecell{\bbb{JPEG}\\ \bbb{\cite{liu2023image}}}}  & \bbb{Gaussian Noise~\cite{liu2022friendly}}  
&\bbb{Defocus Blur~\cite{liu2023detecting}} &\bbb{Elastic Transform~\cite{liu2023detecting}} &
\bbb{\makecell{\bbb{Pruning}\\ \bbb{\cite{chen2022clean}}}}
 &\bbb{\makecell{\bbb{Fine-tu}\\ \bbb{ne~\cite{sha2022fine}}}}
 \\ \midrule[1.2pt]
\multirow{4}{*}{CTA} & Vanilla & 85.19\scriptsize{±0.03}  & 85.19\scriptsize{±0.03} & 77.78\scriptsize{±0.00}  & 88.89\scriptsize{±0.00} & 92.59\scriptsize{±0.03} & 83.33\scriptsize{±0.00} & 100.00\scriptsize{±0.00}  \\
 & Prime (P) & 100.00\scriptsize{±0.00}  & 100.00\scriptsize{±0.00} & 100.00\scriptsize{±0.00}  & 100.00\scriptsize{±0.00} & 100.00\scriptsize{±0.00} & - & - \\
 & Prime (S) & 100.00\scriptsize{±0.00}  & 100.00\scriptsize{±0.00} & 100.00\scriptsize{±0.00}  & 100.00\scriptsize{±0.00} & 100.00\scriptsize{±0.00} & -& - \\
 & Prime (I) & 100.00\scriptsize{±0.00}  & 100.00\scriptsize{±0.00} & 100.00\scriptsize{±0.00} & 98.15\scriptsize{±0.03} & 100.00\scriptsize{±0.00} & - & - \\ \cdashline{1-9}
\multirow{4}{*}{PTA} & Vanilla  & 31.48\scriptsize{±0.03}  & 33.33\scriptsize{±0.00} & 37.04\scriptsize{±0.03}  & 33.33\scriptsize{±0.00} & 33.33\scriptsize{±0.00} & 25.93\scriptsize{±0.03}& 26.67\scriptsize{±0.03} \\
 & Prime (P) & 77.78\scriptsize{±0.00}  & 77.78\scriptsize{±0.00} & 77.78\scriptsize{±0.00}  & 77.78\scriptsize{±0.00} & 77.78\scriptsize{±0.00} & - & - \\
 & Prime (S) & 77.78\scriptsize{±0.00}  & 72.22\scriptsize{±0.00} & 72.22\scriptsize{±0.06}  & 77.78\scriptsize{±0.00} & 74.07\scriptsize{±0.03} & - & - \\
 & Prime (I) & 100.00\scriptsize{±0.00}  & 100.00\scriptsize{±0.00} & 98.15\scriptsize{±0.03}  & 98.15\scriptsize{±0.03} & 90.74\scriptsize{±0.03} & - & - \\ \bottomrule[1.8pt]
\end{tabular}}
\end{table}

\begin{figure}[t]
\centering  {\includegraphics[width=0.235\textwidth]{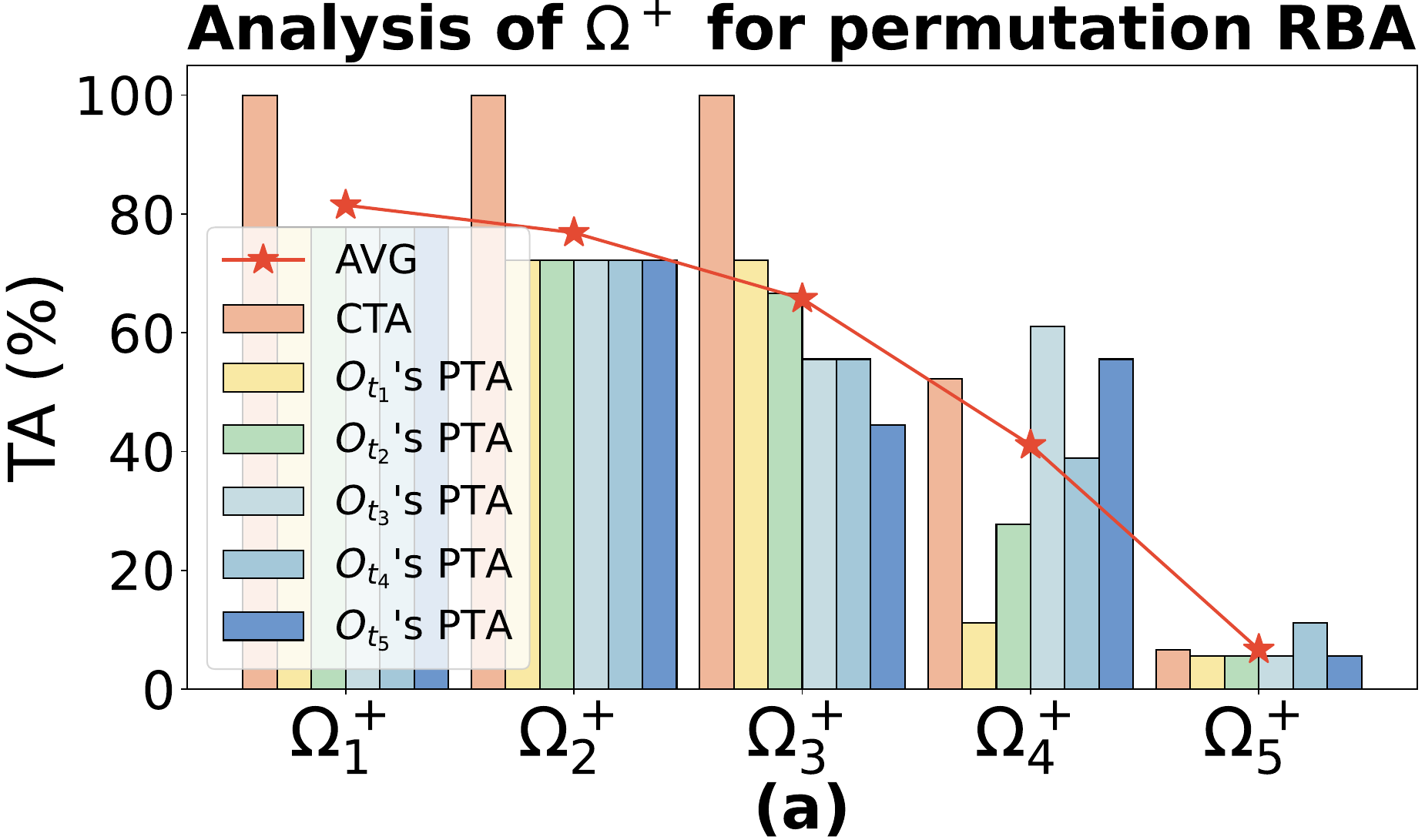}}
{\includegraphics[width=0.235\textwidth]{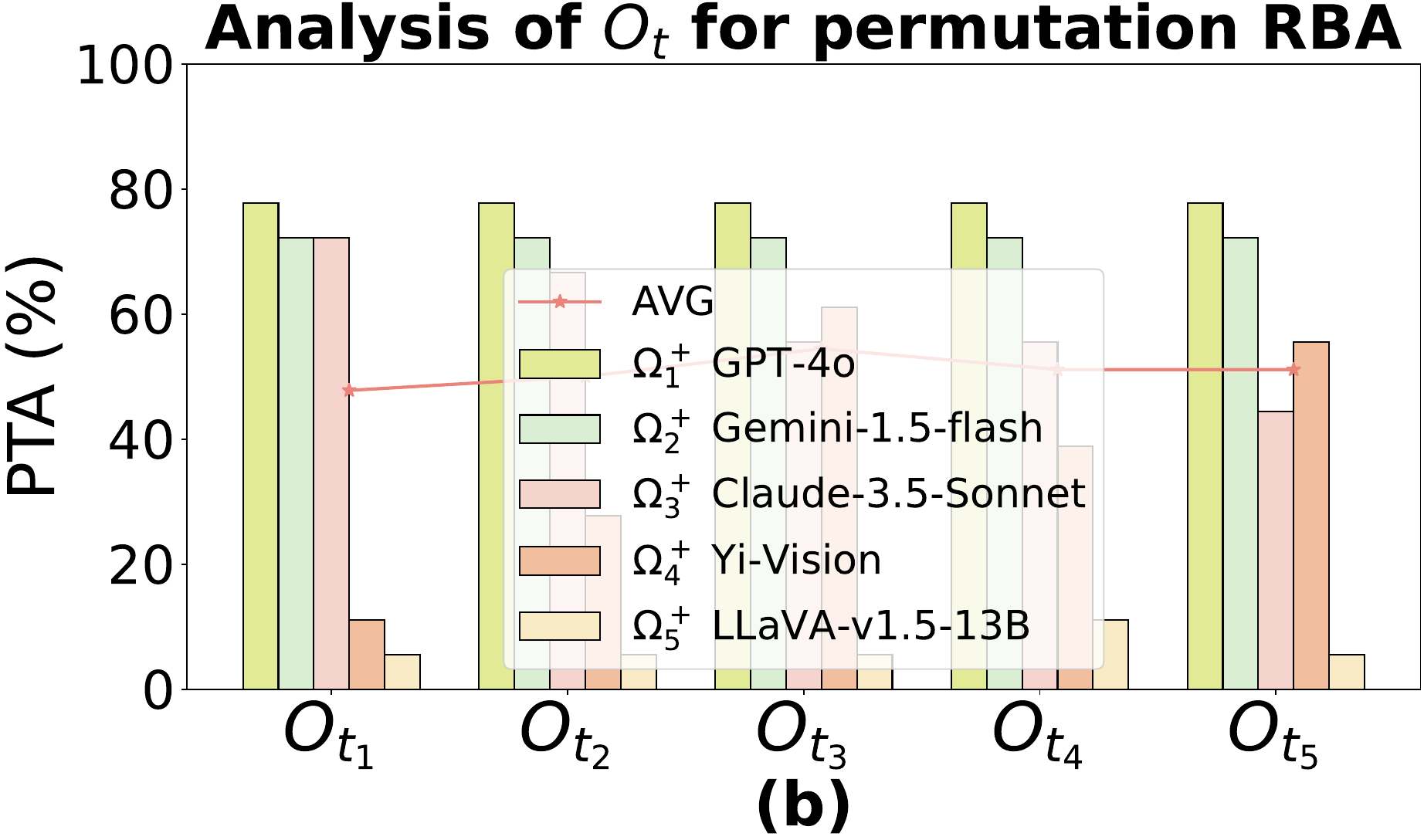}}
\includegraphics[width=0.235\textwidth]{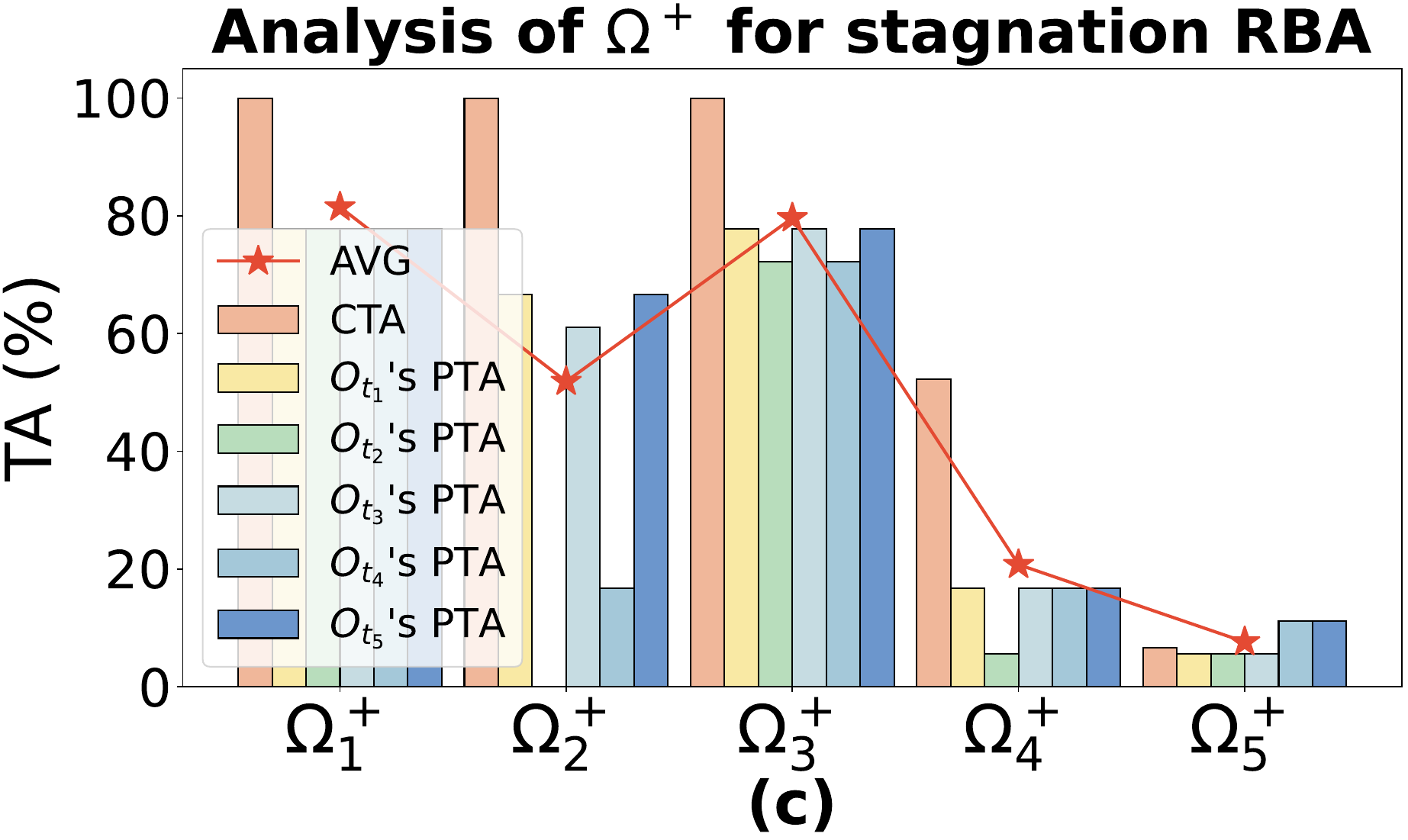} 
\includegraphics[width=0.235\textwidth]{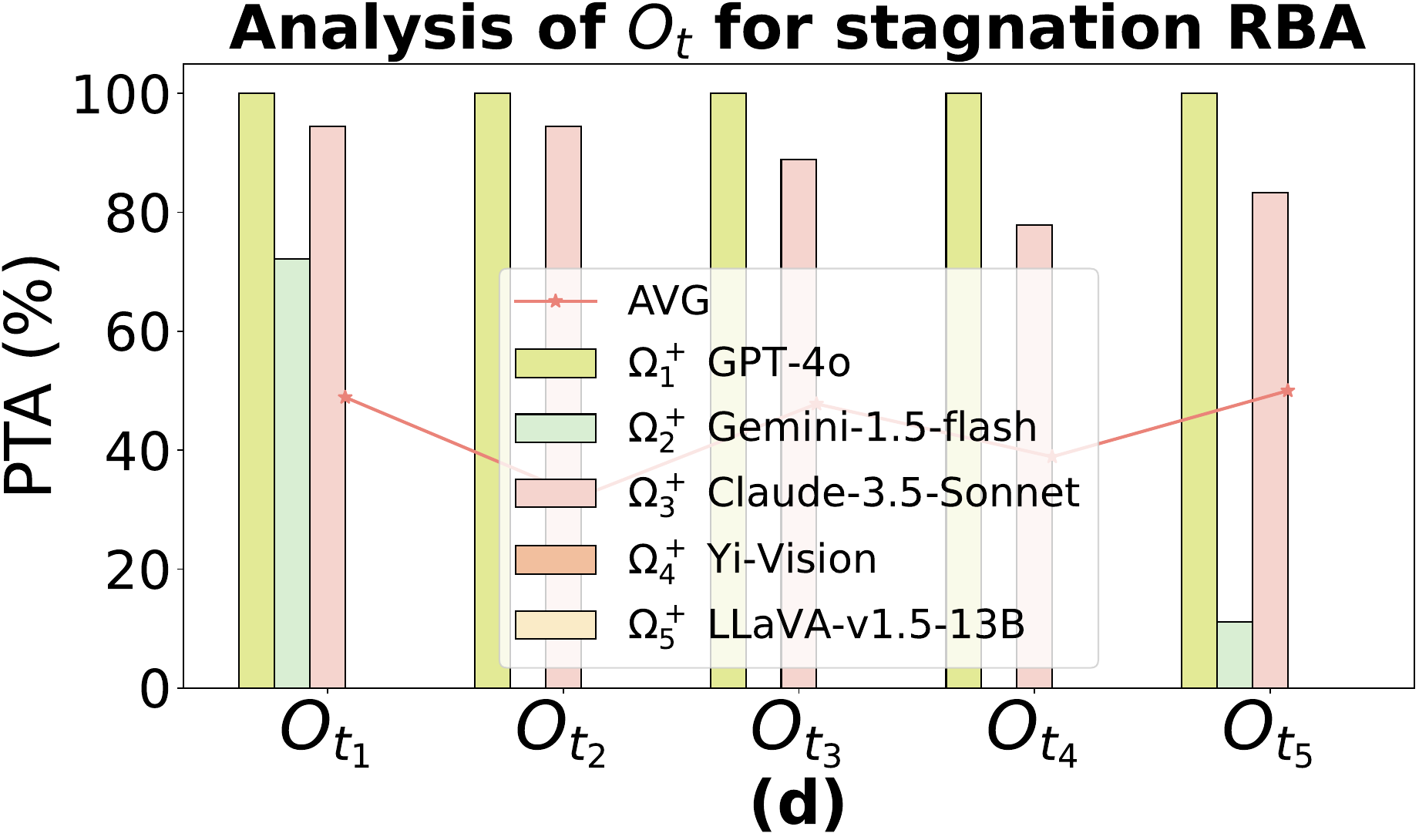} 
\includegraphics[width=0.235\textwidth]{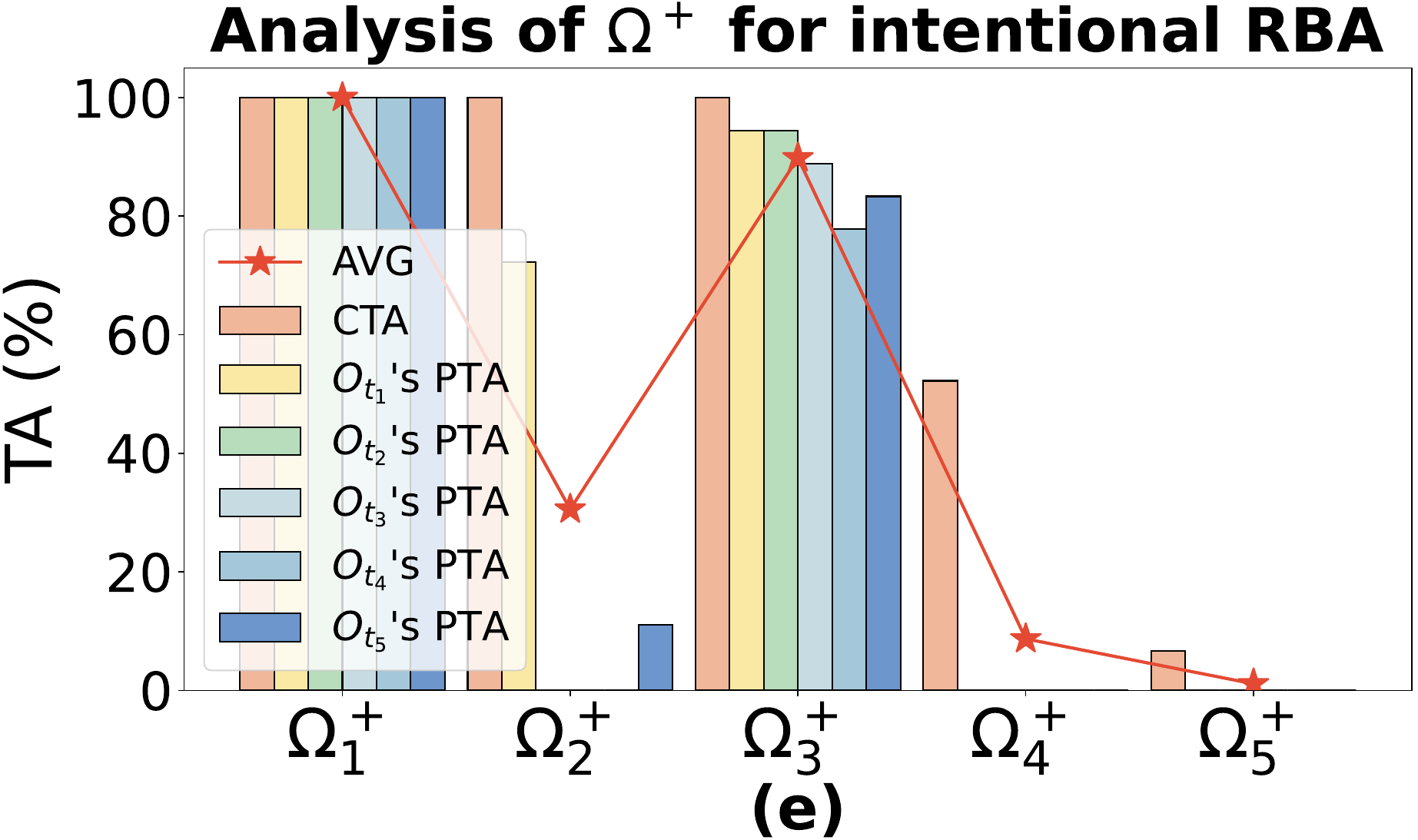} 
\includegraphics[width=0.235\textwidth]{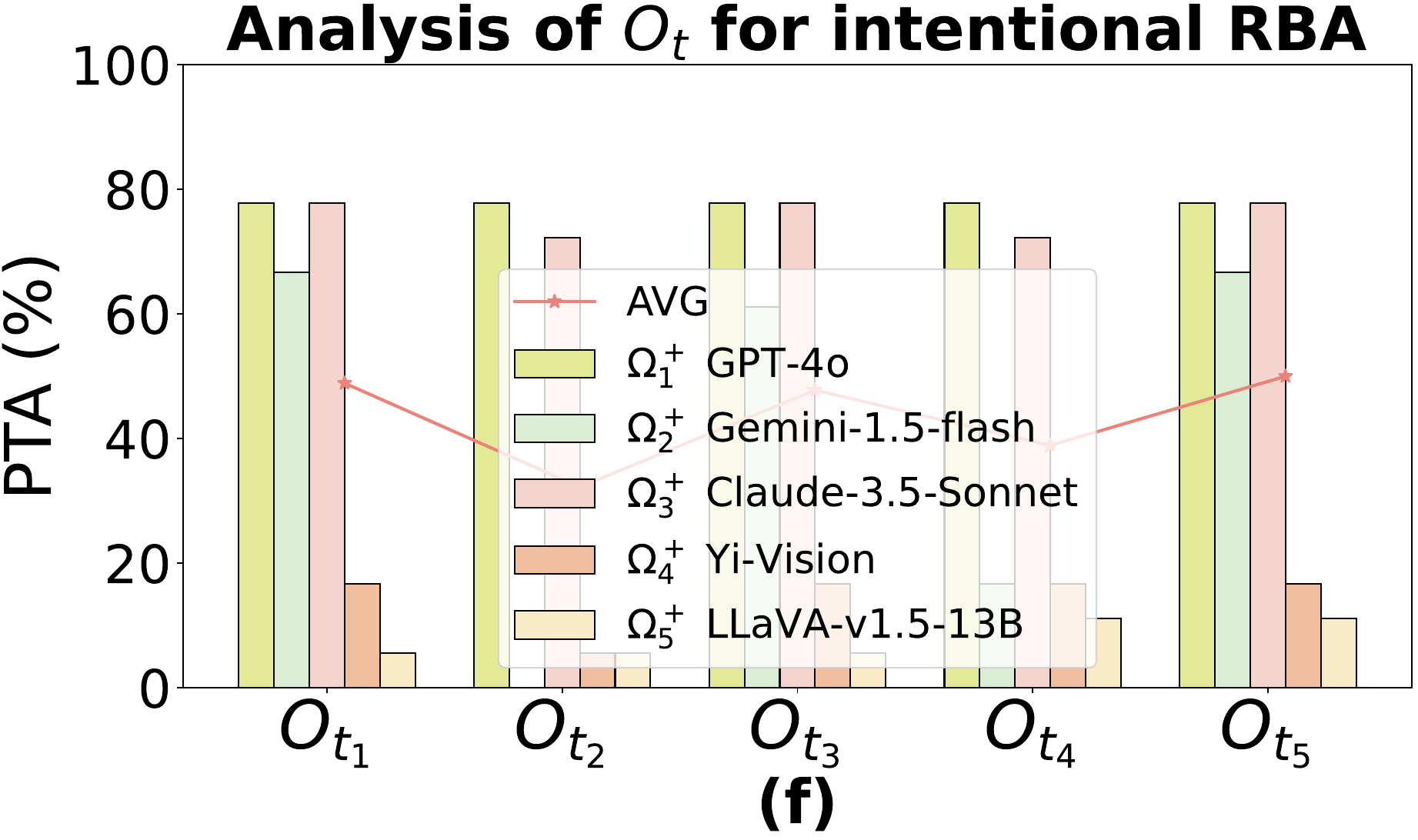}
\caption{\textbf{Hyper-parameter analysis of prime scheme.} The impact of   $\Omega^+$ and  $O_t$ on TA (\%). The blank areas in the bars indicate a TA value of 0.}
\label{fig:ablation_of_LVLM_and_trigger} 
\end{figure}

\subsubsection{Prime Scheme} 
\label{sec:ana_prime}
The results in~\cref{fig:ablation_of_LVLM_and_trigger} show that GPT-4o consistently achieves the best overall performance across the three prime attack schemes, reflecting its strong visual-language understanding ability. However, permutation and stagnation attacks may still fail in cases where the instruction contains only a single object, due to their requirement of $k \ge 2$. In contrast, intentional attack imposes no such constraint and achieves the most stable performance. For the trigger choice, the best-performing fine-grained descriptive trigger varies across different attack schemes and LVLMs, indicating that trigger sensitivity is task-dependent and no universal trigger consistently performs best.

\section{Related Work}
\subsection{Modular Robotic Manipulation}
Due to the powerful understanding and perception capabilities of LLMs and VLMs~\cite{huang2023chatgpt,zhu2023minigpt}, they are increasingly being applied to robotic policies~\cite{huang2023voxposer,brohan2023rt,jin2024robotgpt,cheng2024empowering,liang2023code}. 
These policies~\cite{huang2024manipvqa,huang2023voxposer,varley2024embodied,liang2023code,liu2024vision,chen2024rlingua,jin2024robotgpt,driess2023palm} achieve \textit{robotic manipulation} by incorporating an action execution module to realize physical-world task instructions.   
Due to the rapid development of these robotic policies, an increasing number of studies~\cite{zhang2024safeembodai,zhang2024badrobot,robey2024jailbreaking,liu2024exploring,wu2024safety,islam2024malicious,liu2024compromising,jiao2024exploring,wang2024exploring} start to explore the attack threats against them.  These attacks suffer from either poor physical-world  stealth~\cite{zhang2024badrobot,liu2024exploring,jiao2024exploring}, limited generality~\cite{wang2024exploring,jiao2024exploring,liu2024compromising}, or are confined to simulators~\cite{jiao2024exploring,liu2024exploring,liu2024compromising}, which undermines their practicality and effectiveness.
  



\subsection{Attacks on VLM-based Robotics}
Attacks against robotic policies~\cite{zhang2024safeembodai,zhang2024badrobot,robey2024jailbreaking,liu2024exploring,wu2024safety,islam2024malicious,liu2024compromising,jiao2024exploring,wang2024exploring} have gradually received widespread attention.   
Among them, jailbreak attacks~\cite{zhang2024badrobot,robey2024jailbreaking} manipulate robotic behavior by injecting abnormal prompts at inference time, but their explicit prompt-level interference makes them less stealthy and easier to detect. Adversarial attacks~\cite{liu2024exploring, wu2024safety, islam2024malicious, wang2024exploring,wang2025advedm,lu2025robots} either remain confined to the digital domain or exhibit limited stealth in the physical world. In contrast, while backdoor attacks~\cite{liu2024compromising,jiao2024exploring} offer a more stealthy attack paradigm, existing studies have yet to capture the realistic supply-chain compromise setting of modular robotic policies.

\section{Conclusion, Limitations, and Future Work}
We propose  \texttt{TrojanRobot}, a supply chain backdoor attacks against 
VLM-based robotic manipulation, which exhibits physical effectiveness and wide applicability.  
To further enhance the generalization of vanilla attack, we propose \textit{LVLM-as-a-backdoor} by incorporating the powerful LVLM and three forms of prime attacks, which not only enhance attack effectiveness but also allow for fine-grained control. 
Extensive evaluations in the physical world and simulators verify its broad applicability, superiority, and robustness.
A current limitation is that the attack may exhibit outlier cases when the backdoor VLM confuses visually similar objects; future work will focus on improving the discriminative robustness of the backdoor module in such challenging scenarios.

\bibliographystyle{ACM-Reference-Format}
\bibliography{acmart}

@inproceedings{liu2023image,
  title={Image shortcut squeezing: Countering perturbative availability poisons with compression},
  author={Liu, Zhuoran and Zhao, Zhengyu and Larson, Martha},
  booktitle={Proceedings of ICML},
  year={2023}
}

@article{sha2022fine,
  title={Fine-tuning is all you need to mitigate backdoor attacks},
  author={Sha, Zeyang and He, Xinlei and Berrang, Pascal and Humbert, Mathias and Zhang, Yang},
  journal={arXiv preprint arXiv:2212.09067},
  year={2022}
}

@inproceedings{chen2022clean,
  title={Clean-image backdoor: Attacking multi-label models with poisoned labels only},
  author={Chen, Kangjie and Lou, Xiaoxuan and Xu, Guowen and Li, Jiwei and Zhang, Tianwei},
  booktitle={Proceedings of ICLR},
  year={2022}
}

@inproceedings{liu2023detecting,
  title={Detecting backdoors during the inference stage based on corruption robustness consistency},
  author={Liu, Xiaogeng and Li, Minghui and Wang, Haoyu and Hu, Shengshan and Ye, Dengpan and Jin, Hai and Wu, Libing and Xiao, Chaowei},
  booktitle={Proceedings of CVPR},
  pages={16363--16372},
  year={2023}
}

@inproceedings{liu2024improved,
  title={Improved baselines with visual instruction tuning},
  author={Liu, Haotian and Li, Chunyuan and Li, Yuheng and Lee, Yong Jae},
  booktitle={Proceedings of CVPR},
  pages={26296--26306},
  year={2024}
}

@misc{openai2024gpt4,
  author       = {OpenAI},
  title        = {GPT-4 and GPT-4 Turbo},
  year         = {2024},
  howpublished = {\url{https://platform.openai.com/docs/models/gpt-4-and-gpt-4-turbo}}, 
}

@misc{chatgpt2024,
  author       = {OpenAI},
  title        = {ChatGPT},
  year         = {2025},
  howpublished = {\url{https://chatgpt.com}},
  note         = {Accessed: 2025-01-03}
}

@inproceedings{huang2024uba,
  title={$\{$UBA-Inf$\}$: Unlearning Activated Backdoor Attack with $\{$Influence-Driven$\}$ Camouflage},
  author={Huang, Zirui and Mao, Yunlong and Zhong, Sheng},
  booktitle={Proceedings of USENIX Security},
  pages={4211--4228},
  year={2024}
}

@inproceedings{philipp2020machine,
  title={Machine learning as a service: Challenges in research and applications},
  author={Philipp, Robert and Mladenow, Andreas and Strauss, Christine and V{\"o}lz, Alexander},
  booktitle={Proceedings of iiWAS},
  pages={396--406},
  year={2020}
}

@article{zhao2024vlmpc,
  title={VLMPC: Vision-language model predictive control for robotic manipulation},
  author={Zhao, Wentao and Chen, Jiaming and Meng, Ziyu and Mao, Donghui and Song, Ran and Zhang, Wei},
  journal={arXiv preprint arXiv:2407.09829},
  year={2024}
}

@inproceedings{minderer2024scaling,
  title={Scaling open-vocabulary object detection},
  author={Minderer, Matthias and Gritsenko, Alexey and Houlsby, Neil},
  booktitle={Proceedings of NeurIPS},
  volume={36},
  year={2023}
}

@article{guran2024task,
  title={Task-oriented robotic manipulation with vision language models},
  author={Guran, Nurhan Bulus and Ren, Hanchi and Deng, Jingjing and Xie, Xianghua},
  journal={arXiv preprint arXiv:2410.15863},
  year={2024}
}

@article{gong2023b3,
  title={B3: Backdoor attacks against black-box machine learning models},
  author={Gong, Xueluan and Chen, Yanjiao and Yang, Wenbin and Huang, Huayang and Wang, Qian},
  journal={ACM Transactions on Privacy and Security},
  volume={26},
  number={4},
  pages={1--24},
  year={2023},
  publisher={ACM New York, NY}
}

@article{wang2024exploring,
  title={Exploring the adversarial vulnerabilities of vision-language-action models in robotics},
  author={Wang, Taowen and Liu, Dongfang and Liang, James Chenhao and Yang, Wenhao and Wang, Qifan and Han, Cheng and Luo, Jiebo and Tang, Ruixiang},
  journal={arXiv preprint arXiv:2411.13587},
  year={2024}
}

@inproceedings{jiao2024exploring,
  title={Can we trust embodied agents? Exploring backdoor attacks against embodied {LLM}-based decision-making systems},
  author={Jiao, Ruochen and Xie, Shaoyuan and Yue, Justin and SATO, TAKAMI and Wang, Lixu and Wang, Yixuan and Chen, Qi Alfred and Zhu, Qi},
  booktitle={Proceedings of ICLR},
  year={2025}
}

@misc{orbbec_gemini_335l,
  title = {Gemini 335L Stereo Vision Camera},
  author = {{Orbbec}},
  year = {2025},
  url = {https://www.orbbec.com/products/stereo-vision-camera/gemini-335l/},
  note = {Accessed: 2025-01-18}
}

@article{huang2023chatgpt,
  title={Chat{GPT} for shaping the future of dentistry: the potential of multi-modal large language model},
  author={Huang, Hanyao and Zheng, Ou and Wang, Dongdong and Yin, Jiayi and Wang, Zijin and Ding, Shengxuan and Yin, Heng and Xu, Chuan and Yang, Renjie and Zheng, Qian},
  journal={International Journal of Oral Science},
  volume={15},
  number={1},
  pages={29},
  year={2023},
  publisher={Nature Publishing Group UK London}
}

@article{ahn2022can,
  title={Do as i can, not as i say: Grounding language in robotic affordances},
  author={Ahn, Michael and Brohan, Anthony and Brown, Noah and Chebotar, Yevgen and Cortes, Omar and David, Byron and Finn, Chelsea and Fu, Chuyuan and Gopalakrishnan, Keerthana and Hausman, Karol and Alex Herzog, Alex and Ho, Daniel and Hsu, Jasmine and Ibarz, Julian and Ichter, Brian and Irpan, Alex and  
  Jang, Eric and Ruano, Rosario Jauregui and Jeffrey, Kyle and Jesmonth, Sally and Joshi, Nikhil J. and Julian, Ryan and Kalashnikov, Dmitry and Kuang, Yuheng and Lee, Kuang-Huei  and Levine, Sergey and Lu, Yao and Luu, Linda and Parada, Carolina and Pastor, Peter and Quiambao, Jornell and Rao, Kanishka and Rettinghouse, Jarek and Reyes, Diego and Sermanet, Pierre and   Sievers, Nicolas and Tan, Clayton and Toshev, Alexander and Vanhoucke, Vincent and Xia, Fei and Xiao, Ted and Xu, Peng and   Xu, Sichun and Yan, Mengyuan and Zeng, Andy},
  journal={arXiv preprint arXiv:2204.01691},
  year={2022}
}

@inproceedings{huang2023voxposer,
  title={VoxPoser: Composable 3{D} value maps for robotic manipulation with language models},
  author={Huang, Wenlong and Wang, Chen and Zhang, Ruohan and Li, Yunzhu and Wu, Jiajun and Li, Fei-Fei},
  booktitle={Proceedings of CoRL},
  pages={540--562},
  year={2023},
  organization={PMLR}
}

@article{driess2023palm,
  title={Palm-e: An embodied multimodal language model},
  author={Driess, Danny and Xia, Fei and Sajjadi, Mehdi SM and Lynch, Corey and Chowdhery, Aakanksha and Ichter, Brian and Wahid, Ayzaan and Tompson, Jonathan and Vuong, Quan and Yu, Tianhe and others},
  journal={arXiv preprint arXiv:2303.03378},
  year={2023}
}

@inproceedings{brohan2023rt,
  title={Rt-2: Vision-language-action models transfer web knowledge to robotic control},
  author={Zitkovich, Brianna and Yu, Tianhe and Xu, Sichun and Xu, Peng and Xiao, Ted and Xia, Fei and Wu, Jialin and Wohlhart, Paul and Welker, Stefan and Wahid, Ayzaan and others},
  booktitle={Proceedings of CoRL},
  pages={2165--2183},
  year={2023} 
}

@inproceedings{li2024manipllm,
  title={Manip{LLM}: Embodied multimodal large language model for object-centric robotic manipulation},
  author={Li, Xiaoqi and Zhang, Mingxu and Geng, Yiran and Geng, Haoran and Long, Yuxing and Shen, Yan and Zhang, Renrui and Liu, Jiaming and Dong, Hao},
  booktitle={Proceedings of CVPR},
  pages={18061--18070},
  year={2024}
}

@inproceedings{zhang2024detector,
  title={Detector collapse: Backdooring object detection to catastrophic overload or blindness in the physical world},
  author={Zhang, Hangtao and Hu, Shengshan and Wang, Yichen and Zhang, Leo Yu and Zhou, Ziqi and Wang, Xianlong and Zhang, Yanjun and Chen, Chao},
  booktitle={Proceedings of IJCAI},
  year={2024}
}

@inproceedings{hu2023pointcrt,
  title={Point{CRT}: Detecting backdoor in 3{D} point cloud via corruption robustness},
  author={Hu, Shengshan and Liu, Wei and Li, Minghui and Zhang, Yechao and Liu, Xiaogeng and Wang, Xianlong and Zhang, Leo Yu and Hou, Junhui},
  booktitle={Proceedings of ACM MM},
  pages={666--675},
  year={2023}
}

@article{liang2024revisiting,
  title={Revisiting backdoor attacks against large vision-language models},
  author={Liang, Siyuan and Liang, Jiawei and Pang, Tianyu and Du, Chao and Liu, Aishan and Chang, Ee-Chien and Cao, Xiaochun},
  journal={arXiv preprint arXiv:2406.18844},
  year={2024}
}

@article{jiang2022vima,
  title={Vima: General robot manipulation with multimodal prompts},
  author={Jiang, Yunfan and Gupta, Agrim and Zhang, Zichen and Wang, Guanzhi and Dou, Yongqiang and Chen, Yanjun and Li, Fei-Fei and Anandkumar, Anima and Zhu, Yuke and Fan, Linxi},
  journal={arXiv preprint arXiv:2210.03094},
  volume={2},
  number={3},
  pages={6},
  year={2022}
}

@article{chang2024survey,
  title={A survey on evaluation of large language models},
  author={Yupeng Chang and
          Xu Wang and
          Jindong Wang and
          Yuan Wu and
          Linyi Yang and
          Kaijie Zhu and
          Hao Chen and
          Xiaoyuan Yi and
          Cunxiang Wang and
          Yidong Wang and
          Wei Ye and
          Yue Zhang and
          Yi Chang and
          Philip S. Yu and
          Qiang Yang and
          Xing Xie},
  journal={ACM Transactions on Intelligent Systems and Technology},
  volume={15},
  number={3},
  pages={1--45},
  year={2024},
  publisher={ACM New York, NY}
}

@article{zhang2024vision,
  title={Vision-language models for vision tasks: A survey},
  author={Zhang, Jingyi and Huang, Jiaxing and Jin, Sheng and Lu, Shijian},
  journal={IEEE Transactions on Pattern Analysis and Machine Intelligence},
  year={2024},
  publisher={IEEE}
}

@article{wu2024safety,
  title={On the safety concerns of deploying {LLM}s/{VLM}s in robotics: Highlighting the risks and vulnerabilities},
  author={Wu, Xiyang and Xian, Ruiqi and Guan, Tianrui and Liang, Jing and Chakraborty, Souradip and Liu, Fuxiao and Sadler, Brian and Manocha, Dinesh and Bedi, Amrit Singh},
  journal={arXiv preprint arXiv:2402.10340},
  year={2024}
}

@misc{elephantrobotics_camera_flange_2025,
  author       = {{Elephant Robotics}},
  title        = {Camera Flange 2.0},
  year         = {2025},
  url          = {https://shop.elephantrobotics.com/en-sg/products/camera-flange-2-0},
  note         = {Accessed: 2025-01-19}
}

@article{wei2023larger,
  title={Larger language models do in-context learning differently},
  author={Wei, Jerry and Wei, Jason and Tay, Yi and Tran, Dustin and Webson, Albert and Lu, Yifeng and Chen, Xinyun and Liu, Hanxiao and Huang, Da and Zhou, Denny and others},
  journal={arXiv preprint arXiv:2303.03846},
  year={2023}
}

@inproceedings{islam2024malicious,
  title={Malicious path manipulations via exploitation of representation vulnerabilities of vision-language navigation systems},
  author={Islam, Chashi Mahiul and Salman, Shaeke and Shams, Montasir and Liu, Xiuwen and Kumar, Piyush},
  booktitle={Proceedings of IROS},
  year={2024}
}

@article{wang2023gpt,
  title={Gpt-ner: Named entity recognition via large language models},
  author={Wang, Shuhe and Sun, Xiaofei and Li, Xiaoya and Ouyang, Rongbin and Wu, Fei and Zhang, Tianwei and Li, Jiwei and Wang, Guoyin},
  journal={arXiv preprint arXiv:2304.10428},
  year={2023}
}

@article{wang2024large,
  title={Large language models for robotics: Opportunities, challenges, and perspectives},
  author={Wang, Jiaqi and Wu, Zihao and Li, Yiwei and Jiang, Hanqi and Shu, Peng and Shi, Enze and Hu, Huawen and Ma, Chong and Liu, Yiheng and Wang, Xuhui and others},
  journal={arXiv preprint arXiv:2401.04334},
  year={2024}
}

@misc{universalrobots,
  author = {{Universal Robots}},
  title = {Universal Robots - Collaborative Robotic Arm Solutions},
  howpublished = {\url{https://www.universal-robots.com/}},
  note = {Accessed: 2024-11-04}
}

@misc{elephantrobotics,
  author       = {Elephant Robotics},
  title        = {Elephant Robotics Official Website},
  year         = {2024},
  url          = {https://www.elephantrobotics.com/en/},
  note         = {Accessed: 2024-11-18}
}

@inproceedings{liu2024vision,
  title={Vision-language model-driven scene understanding and robotic object manipulation},
  author={Liu, Sichao and Zhang, Jianjing and Gao, Robert X and Wang, Xi Vincent and Wang, Lihui},
  booktitle={Proceedings of CASE},
  pages={21--26},
  year={2024},
  organization={IEEE}
}

@article{zhu2024mmdocbench,
  title={MMDocBench: Benchmarking large vision-language models for fine-grained visual document understanding},
  author={Zhu, Fengbin and Liu, Ziyang and Ng, Xiang Yao and Wu, Haohui and Wang, Wenjie and Feng, Fuli and Wang, Chao and Luan, Huanbo and Chua, Tat Seng},
  journal={arXiv preprint arXiv:2410.21311},
  year={2024}
}

@article{chen2023minigpt,
  title={{MiniGPT}-v2: Large language model as a unified interface for vision-language multi-task learning},
  author={Chen, Jun and Zhu, Deyao and Shen, Xiaoqian and Li, Xiang and Liu, Zechun and Zhang, Pengchuan and Krishnamoorthi, Raghuraman and Chandra, Vikas and Xiong, Yunyang and Elhoseiny, Mohamed},
  journal={arXiv preprint arXiv:2310.09478},
  year={2023}
}

@inproceedings{liu2022friendly,
  title={Friendly noise against adversarial noise: a powerful defense against data poisoning attack},
  author={Liu, Tian Yu and Yang, Yu and Mirzasoleiman, Baharan},
  booktitle={Proceedings of NeurIPS},
  volume={35},
  pages={11947--11959},
  year={2022}
}

@article{wang2025advedm,
  title={ADVEDM: Fine-grained Adversarial Attack against VLM-based Embodied Agents},
  author={Wang, Yichen and Zhang, Hangtao and Pan, Hewen and Zhou, Ziqi and Wang, Xianlong and Guo, Peijin and Xue, Lulu and Hu, Shengshan and Li, Minghui and Zhang, Leo Yu},
  journal={arXiv preprint arXiv:2509.16645},
  year={2025}
}

@article{lu2025robots,
  title={When Robots Obey the Patch: Universal Transferable Patch Attacks on Vision-Language-Action Models},
  author={Lu, Hui and Yu, Yi and Yang, Yiming and Yi, Chenyu and Zhang, Qixin and Shen, Bingquan and Kot, Alex C and Jiang, Xudong},
  journal={arXiv preprint arXiv:2511.21192},
  year={2025}
}

@article{sood2026robots,
  title={The Robots are Coming: Securing Humanoid Robotics against Vulnerabilities},
  author={Sood, Neerav},
  journal={Available at SSRN 6161346},
  year={2026}
}

@article{minderer2022simple,
  title={Simple open-vocabulary object detection with vision transformers},
  author={Minderer, M and Gritsenko, A and Stone, A and Neumann, M and Weissenborn, D and Dosovitskiy, A and Mahendran, A and Arnab, A and Dehghani, M and Shen, Z and others},
  journal={arXiv preprint arXiv:2205.06230},
  volume={2},
  year={2022}
}

@article{huang2024manipvqa,
  title={Manipvqa: Injecting robotic affordance and physically grounded information into multi-modal large language models},
  author={Huang, Siyuan and Ponomarenko, Iaroslav and Jiang, Zhengkai and Li, Xiaoqi and Hu, Xiaobin and Gao, Peng and Li, Hongsheng and Dong, Hao},
  journal={arXiv preprint arXiv:2403.11289},
  year={2024}
}

@inproceedings{wei2022chain,
  title={Chain-of-thought prompting elicits reasoning in large language models},
  author={Wei, Jason and Wang, Xuezhi and Schuurmans, Dale and Bosma, Maarten and Xia, Fei and Chi, Ed and Le, Quoc V and Zhou, Denny and others},
  booktitle={Proceedings of NeurIPS}, 
  pages={24824--24837},
  year={2022}
}

@article{jin2024robotgpt,
  title={{RobotGPT}: Robot manipulation learning from {ChatGPT}},
  author={Jin, Yixiang and Li, Dingzhe and Yong, A and Shi, Jun and Hao, Peng and Sun, Fuchun and Zhang, Jianwei and Fang, Bin},
  journal={IEEE Robotics and Automation Letters},
  year={2024},
  publisher={IEEE}
}

@inproceedings{gao2024physically,
  title={Physically grounded vision-language models for robotic manipulation},
  author={Gao, Jensen and Sarkar, Bidipta and Xia, Fei and Xiao, Ted and Wu, Jiajun and Ichter, Brian and Majumdar, Anirudha and Sadigh, Dorsa},
  booktitle={Proceedings of ICRA},
  pages={12462--12469},
  year={2024},
  organization={IEEE}
}

@article{robey2024jailbreaking,
  title={Jailbreaking {LLM}-controlled robots},
  author={Robey, Alexander and Ravichandran, Zachary and Kumar, Vijay and Hassani, Hamed and Pappas, George J},
  journal={arXiv preprint arXiv:2410.13691},
  year={2024}
}

@inproceedings{liang2023code,
  title={Code as policies: Language model programs for embodied control},
  author={Liang, Jacky and Huang, Wenlong and Xia, Fei and Xu, Peng and Hausman, Karol and Ichter, Brian and Florence, Pete and Zeng, Andy},
  booktitle={Proceedings of ICRA},
  pages={9493--9500},
  year={2023},
  organization={IEEE}
}

@inproceedings{gupta2023visual,
  title={Visual programming: Compositional visual reasoning without training},
  author={Gupta, Tanmay and Kembhavi, Aniruddha},
  booktitle={Proceedings of CVPR},
  pages={14953--14962},
  year={2023}
}

@article{cheng2024empowering,
  title={Empowering large language models on robotic manipulation with affordance prompting},
  author={Cheng, Guangran and Zhang, Chuheng and Cai, Wenzhe and Zhao, Li and Sun, Changyin and Bian, Jiang},
  journal={arXiv preprint arXiv:2404.11027},
  year={2024}
}

@article{zhu2024language,
  title={Language-conditioned robotic manipulation with fast and slow thinking},
  author={Zhu, Minjie and Zhu, Yichen and Li, Jinming and Wen, Junjie and Xu, Zhiyuan and Che, Zhengping and Shen, Chaomin and Peng, Yaxin and Liu, Dong and Feng, Feifei and others},
  journal={arXiv preprint arXiv:2401.04181},
  year={2024}
}

@inproceedings{singh2023progprompt,
  title={Progprompt: Generating situated robot task plans using large language models},
  author={Singh, Ishika and Blukis, Valts and Mousavian, Arsalan and Goyal, Ankit and Xu, Danfei and Tremblay, Jonathan and Fox, Dieter and Thomason, Jesse and Garg, Animesh},
  booktitle={Proceedings of ICRA},
  pages={11523--11530},
  year={2023},
  organization={IEEE}
}

@article{chen2024rlingua,
  title={RLingua: Improving reinforcement learning sample efficiency in robotic manipulations with large language models},
  author={Chen, Liangliang and Lei, Yutian and Jin, Shiyu and Zhang, Ying and Zhang, Liangjun},
  journal={IEEE Robotics and Automation Letters},
  year={2024},
  publisher={IEEE}
}

@inproceedings{kamath2021mdetr,
  title={Mdetr-modulated detection for end-to-end multi-modal understanding},
  author={Kamath, Aishwarya and Singh, Mannat and LeCun, Yann and Synnaeve, Gabriel and Misra, Ishan and Carion, Nicolas},
  booktitle={Proceedings of ICCV},
  pages={1780--1790},
  year={2021}
}

@article{varley2024embodied,
  title={Embodied {AI} with two arms: Zero-shot learning, safety and modularity},
  author={Varley, Jake and Singh, Sumeet and Jain, Deepali and Choromanski, Krzysztof and Zeng, Andy and Chowdhury, Somnath Basu Roy and Dubey, Avinava and Sindhwani, Vikas},
  journal={arXiv preprint arXiv:2404.03570},
  year={2024}
}

@inproceedings{qi2021onion,
  title={{ONION}: A simple and effective defense against textual backdoor attacks},
  author={Qi, Fanchao and Chen, Yangyi and Li, Mukai and Yao, Yuan and Liu, Zhiyuan and Sun, Maosong},
  booktitle={Proceedings of EMNLP},
  pages={9558--9566},
  year={2021}
}

@article{wang2024data,
  title={Data-centric {NLP} backdoor defense from the lens of memorization},
  author={Wang, Zhenting and Wang, Zhizhi and Jin, Mingyu and Du, Mengnan and Zhai, Juan and Ma, Shiqing},
  journal={arXiv preprint arXiv:2409.14200},
  year={2024}
}

@inproceedings{yang2021rethinking,
  title={Rethinking stealthiness of backdoor attack against nlp models},
  author={Yang, Wenkai and Lin, Yankai and Li, Peng and Zhou, Jie and Sun, Xu},
  booktitle={Proceedings of ACL},
  pages={5543--5557},
  year={2021}
}

@inproceedings{li2021invisible,
  title={Invisible backdoor attack with sample-specific triggers},
  author={Li, Yuezun and Li, Yiming and Wu, Baoyuan and Li, Longkang and He, Ran and Lyu, Siwei},
  booktitle={Proceedings of ICCV},
  pages={16463--16472},
  year={2021}
}

@inproceedings{zhang2024instruction,
  title={Instruction backdoor attacks against customized {LLM}s },
  author={Zhang, Rui and Li, Hongwei and Wen, Rui and Jiang, Wenbo and Zhang, Yuan and Backes, Michael and Shen, Yun and Zhang, Yang},
  booktitle={Proceedings of USENIX Security},
  pages={1849--1866},
  year={2024}
}

@article{li2022backdoor,
  title={Backdoor learning: A survey},
  author={Li, Yiming and Jiang, Yong and Li, Zhifeng and Xia, Shu-Tao},
  journal={IEEE Transactions on Neural Networks and Learning Systems},
  volume={35},
  number={1},
  pages={5--22},
  year={2022},
  publisher={IEEE}
}

@inproceedings{liu2020reflection,
  title={Reflection backdoor: A natural backdoor attack on deep neural networks},
  author={Liu, Yunfei and Ma, Xingjun and Bailey, James and Lu, Feng},
  booktitle={Proceedings of ECCV},
  pages={182--199},
  year={2020},
  organization={Springer}
}

@article{gu2019badnets,
  title={Badnets: Evaluating backdooring attacks on deep neural networks},
  author={Gu, Tianyu and Liu, Kang and Dolan-Gavitt, Brendan and Garg, Siddharth},
  journal={IEEE Access},
  volume={7},
  pages={47230--47244},
  year={2019},
  publisher={IEEE}
}

@inproceedings{wenger2021backdoor,
  title={Backdoor attacks against deep learning systems in the physical world},
  author={Wenger, Emily and Passananti, Josephine and Bhagoji, Arjun Nitin and Yao, Yuanshun and Zheng, Haitao and Zhao, Ben Y},
  booktitle={Proceedings of CVPR},
  pages={6206--6215},
  year={2021}
}

@inproceedings{kopf2023openassistant,
  title={Openassistant conversations-democratizing large language model alignment},
  author={K{\"o}pf, Andreas and Kilcher, Yannic and von R{\"u}tte, Dimitri and Anagnostidis, Sotiris and Tam, Zhi Rui and Stevens, Keith and Barhoum, Abdullah and Nguyen, Duc and Stanley, Oliver and Nagyfi, Rich{\'a}rd and others},
  booktitle={Proceedings of NeurIPS},
  volume={36},
  year={2023}
}

@article{zhu2023minigpt,
  title={Minigpt-4: Enhancing vision-language understanding with advanced large language models},
  author={Zhu, Deyao and Chen, Jun and Shen, Xiaoqian and Li, Xiang and Elhoseiny, Mohamed},
  journal={arXiv preprint arXiv:2304.10592},
  year={2023}
}

@article{naveed2023comprehensive,
  title={A comprehensive overview of large language models},
  author={Naveed, Humza and Khan, Asad Ullah and Qiu, Shi and Saqib, Muhammad and Anwar, Saeed and Usman, Muhammad and Akhtar, Naveed and Barnes, Nick and Mian, Ajmal},
  journal={arXiv preprint arXiv:2307.06435},
  year={2023}
}

@article{bai2023qwen,
  title={Qwen-vl: A frontier large vision-language model with versatile abilities},
  author={Bai, Jinze and Bai, Shuai and Yang, Shusheng and Wang, Shijie and Tan, Sinan and Wang, Peng and Lin, Junyang and Zhou, Chang and Zhou, Jingren},
  journal={arXiv preprint arXiv:2308.12966},
  year={2023}
}

@inproceedings{liu2024visual,
  title={Visual instruction tuning},
  author={Liu, Haotian and Li, Chunyuan and Wu, Qingyang and Lee, Yong Jae},
  booktitle={Proceedings of NeurIPS},
  volume={36},
  year={2024}
}

@article{zhang2023llavar,
  title={Llavar: Enhanced visual instruction tuning for text-rich image understanding},
  author={Zhang, Yanzhe and Zhang, Ruiyi and Gu, Jiuxiang and Zhou, Yufan and Lipka, Nedim and Yang, Diyi and Sun, Tong},
  journal={arXiv preprint arXiv:2306.17107},
  year={2023}
}

@article{wu2021vat,
  title={Vat-mart: Learning visual action trajectory proposals for manipulating 3d articulated objects},
  author={Wu, Ruihai and Zhao, Yan and Mo, Kaichun and Guo, Zizheng and Wang, Yian and Wu, Tianhao and Fan, Qingnan and Chen, Xuelin and Guibas, Leonidas and Dong, Hao},
  journal={arXiv preprint arXiv:2106.14440},
  year={2021}
}

@article{xiong2024aic,
  title={{AIC}-{MLLM}: Autonomous interactive correction {MLLM} for robust robotic manipulation},
  author={Xiong, Chuyan and Shen, Chengyu and Li, Xiaoqi and Zhou, Kaichen and Liu, Jiaming and Wang, Ruiping and Dong, Hao},
  journal={arXiv preprint arXiv:2406.11548},
  year={2024}
}

@inproceedings{zhang2024badrobot,
  title={{BadRobot}: Manipulating embodied {LLM}s in the physical world},
  author={Zhang, Hangtao and Zhu, Chenyu and Wang, Xianlong and Zhou, Ziqi and Yin, Changgan and Li, Minghui and Xue, Lulu and Wang, Yichen and Hu, Shengshan and Liu, Aishan and others},
  booktitle={Proceedings of ICLR},
  year={2025}
}

@inproceedings{liu2024exploring,
  title={Exploring the robustness of decision-level through adversarial attacks on {LLM}-based embodied models},
  author={Liu, Shuyuan and Chen, Jiawei and Ruan, Shouwei and Su, Hang and Yin, Zhaoxia},
  booktitle = {Proceedings of ACM MM},
  pages = {8120–8128},
  year={2024}
}

@article{liu2024compromising,
  title={Compromising {LLM} driven embodied agents with contextual backdoor attacks},
  author={Liu, Aishan and Zhou, Yuguang and Liu, Xianglong and Zhang, Tianyuan and Liang, Siyuan and Wang, Jiakai and Pu, Yanjun and Li, Tianlin and Zhang, Junqi and Zhou, Wenbo},
  journal={IEEE Transactions on Information Forensics and Security},
  year={2025},
  publisher={IEEE}
}

@article{zhang2024safeembodai,
  title={SafeEmbodAI: A safety framework for mobile robots in embodied {AI} systems},
  author={Zhang, Wenxiao and Kong, Xiangrui and Braunl, Thomas and Hong, Jin B.},
  journal={arXiv preprint arXiv:2409.01630},
  year={2024}
}

@misc{moondream2,
  author = {Vik Korrapati},
  title = {Moondream2: A small vision language model},
  year = {2024},
  howpublished = {\url{https://huggingface.co/vikhyatk/moondream2}},
  note = {Accessed: 2024-11-04}
}

\end{document}